\documentclass[runningheads]{llncs}

\usepackage[mobile]{eccv}

\usepackage{eccvabbrv}

\usepackage{graphicx}
\usepackage{booktabs}

\usepackage[accsupp]{axessibility}  

\definecolor{alizarin}{rgb}{0.82, 0.1, 0.26}
\usepackage[colorlinks=true, urlcolor=alizarin]{hyperref}

\usepackage{orcidlink}

\usepackage{graphicx}
\usepackage{amsmath}
\usepackage{amssymb}
\usepackage{booktabs}

\usepackage[ruled,vlined]{algorithm2e}
\usepackage{cuted}
\usepackage{fvextra}   

\usepackage{booktabs}
\usepackage{makecell} 
\usepackage{pifont}   
\usepackage{multirow}
\usepackage{xcolor}
\newcommand{\cmark}{\ding{51}}
\newcommand{\xmark}{\ding{55}}

\usepackage{graphicx}
\usepackage{cuted}     
\usepackage{capt-of}

\usepackage{tcolorbox}
\tcbuselibrary{listings,skins,breakable}
\usepackage{enumitem}

\usepackage{caption}
\usepackage{cuted}
\usepackage{enumitem}
\usepackage{tabularx}

\newcommand{\mypara}[1]{\paragraph{\textbf{#1}}}

\begin{document}

\title{LangDriveCTRL: Natural Language Controllable Driving Scene Editing with Multi-modal Agents} 

\titlerunning{LangDriveCTRL}

\author{Yun He\inst{1} \and
Francesco Pittaluga\inst{2} \and Ziyu Jiang\inst{2} \and Matthias Zwicker\inst{1}  \and \\ Manmohan Chandraker\inst{2,3} \and
Zaid Tasneem\inst{2}}

\authorrunning{Y. He~et al.}

\institute{University of Maryland, College Park, USA  \and
NEC Labs America, USA \and
UC San Diego, USA \\
{\hypersetup{urlcolor=alizarin}\url{https://yunhe24.github.io/langdrivectrl/}}
}

\maketitle

\vspace{-1.5em}

\begin{figure}[ht]
    \centering
    \begin{minipage}{0.96\linewidth}
        \centering
        \includegraphics[width=\linewidth]{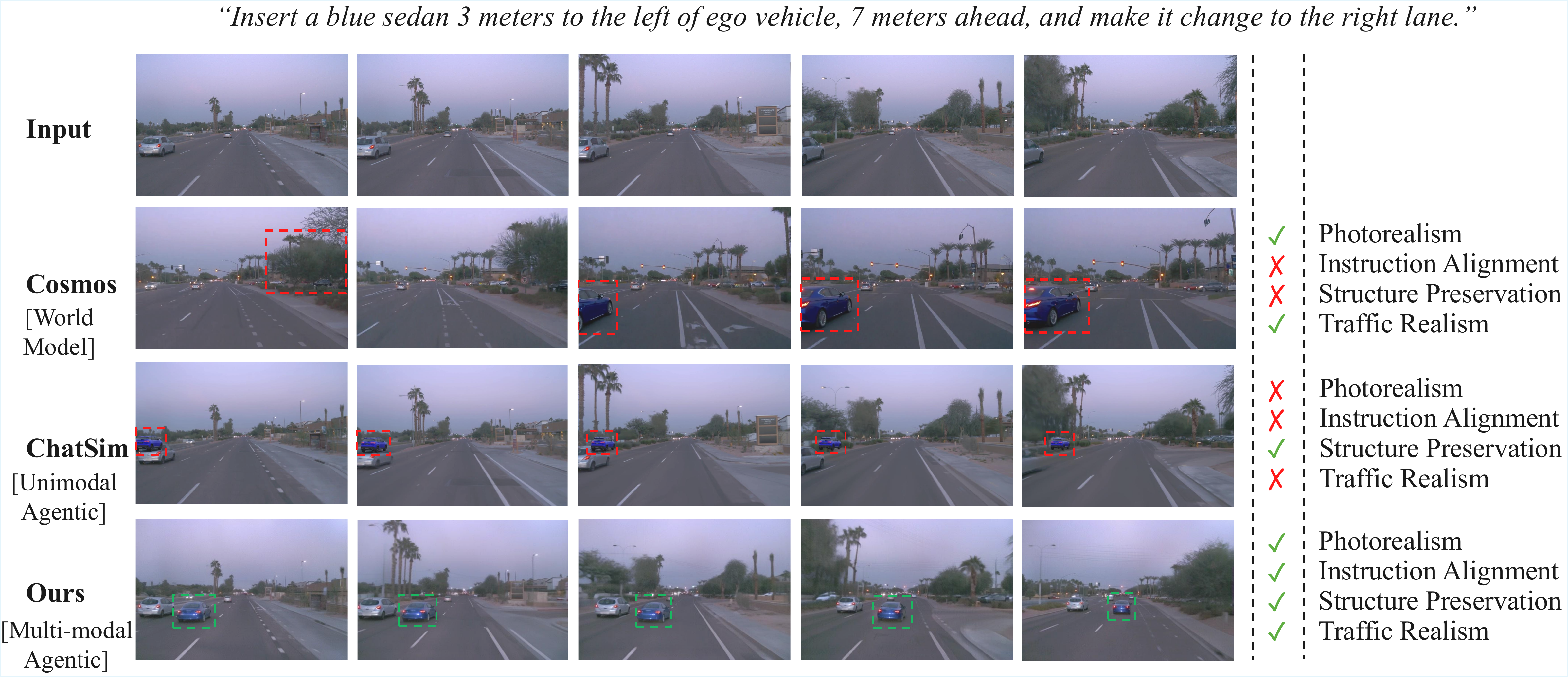}
        \captionof{figure}{\textbf{Natural-language editing.} Cosmos \cite{ali2025world} achieves high visual quality but fails to align with the target behavior and modifies the background, showing poor controllability.
        While ChatSim \cite{wei2024editable} preserves background information, it suffers from poor photorealism, inaccurate trajectory generation, and traffic violation (e.g., collision).
        In contrast, our method achieves photorealism, instruction alignment, structure preservation, and traffic realism simultaneously, significantly outperforming previous methods.}
        \label{fig:teaser}
    \end{minipage}
\end{figure}

\vspace{-2.5em}

\begin{abstract}
LangDriveCTRL is a natural-language-controllable framework for editing real-world driving videos to synthesize diverse traffic scenarios. 
It represents each video as an explicit 3D scene graph, decomposing the scene into a static background and dynamic object nodes.
To enable fine-grained editing and realism, it introduces a \textit{feedback-driven} agentic pipeline. 
An \textit{Orchestrator} converts user instructions into executable graphs that coordinate specialized multi-modal agents and tools.
An \textit{Object Grounding Agent} aligns free-form text with target object nodes in the scene graph; a \textit{Behavior Editing Agent} generates multi-object trajectories from language instructions; and a \textit{Behavior Reviewer Agent} iteratively reviews and refines the generated trajectories. 
The edited scene graph is rendered and harmonized using a video diffusion tool, and then further refined by a \textit{Video Reviewer Agent} to ensure photorealism and appearance alignment.
LangDriveCTRL supports both object node editing (removal, insertion, and replacement) and multi-object behavior editing from natural-language instructions. 
Quantitatively, it achieves nearly $2\times$ higher instruction alignment than the previous SoTA, with superior photorealism, structural preservation, and traffic realism.

\keywords{Video Editing \and Multi-modal Agents \and Autonomous Driving}
\end{abstract}

\section{Introduction}
Synthetic data generation \cite{song2023synthetic} is increasingly adopted to address the limited diversity and coverage of real-world driving logs \cite{ieeeSelfDrivingCars}, especially for training and validating autonomous driving stacks. 
Because collecting real driving videos, particularly those depicting safety-critical scenarios, is prohibitively expensive and logistically impractical \cite{xu2025wod}. 
Traditional driving simulators such as CARLA \cite{dosovitskiy2017carla} and AirSim \cite{shah2017airsim} can generate diverse scenarios.
However, they rely on manually created 3D assets and require engineers to write scripts for scenario generation and human feedback for refinement.

Recent works attempt to scale these workflows by enabling natural-language-driven scene editing. Agentic pipelines~\cite{hsu2025autovfx,wei2024chatdyn,wei2024editable} leverage explicit 3D representations and Large Language Models (LLMs)~\cite{achiam2023gpt} to orchestrate modular tools. 
However, they suffer from three key issues. 
1) They rely solely on unimodal text reasoning without integrating multimodal scene context, which makes it difficult to accurately localize the target objects and generate realistic trajectories.
2) They simply composite the background with the inserted object, resulting in poor rendering quality under large viewpoint changes and failing to achieve lighting-aware insertion. 
3) Most importantly, they do not verify intermediate results after each step, leading to error accumulation and poor final results.

In contrast, implicit world models such as Cosmos~\cite{ali2025world} directly edit videos in pixel space rather than 3D space, achieving strong photorealism and plausible object behavior.
However, it sacrifices controllability. Specifically, it does not explicitly support object-level editing and may unintentionally alter scene structure (e.g., inserting unrequested objects). Like agentic pipelines, it is also feed-forward, lacking feedback mechanisms to correct instruction misalignment.

To address these challenges, we propose \textbf{LangDriveCTRL}, a feedback-driven, natural-language-controllable framework that unifies explicit scene representation with diffusion-based behavior and video refinement. 
Our approach is based on two key insights.
1) Fine-grained controllability requires \emph{multi-modal reasoning} that jointly grounds language instructions in visual appearance and traffic context.
2) Photorealism and instruction alignment require \emph{feedback-driven iterative refinement}, where intermediate behaviors and renderings are reviewed and corrected in a closed loop.

LangDriveCTRL operates on a scene-graph representation obtained via explicit 3D decomposition. 
Each video is modeled as a static background node and dynamic object nodes with trajectories.
This design enables object-level editing while preserving scene structure. 
A central LLM-based \emph{Orchestrator} coordinates reasoning-capable \emph{agents} and functional \emph{tools}. 
Agents (driven by LLMs or VLMs) interpret user intent, ground language instructions in scene context, reason about traffic semantics, and iteratively review and refine intermediate outputs.
While tools execute atomic operations such as 3D reconstruction~\cite{kerbl20233d,chen2024omnire}, text-to-3D generation~\cite{zhao2025hunyuan3d}, and multi-object trajectory simulation~\cite{chang2025langtraj}.

Given a user instruction, the \emph{Orchestrator} first decomposes it into object-level sub-tasks and constructs an execution workflow.  
An \emph{Object Grounding Agent} matches open-vocabulary descriptions to object nodes by jointly reasoning over appearance, behavior, and position information.
For behavior editing, a \emph{Behavior Editing Agent} generates counterfactual behavior based on trajectory history and lane information, and invokes a diffusion-based multi-object simulator \cite{chang2025langtraj} to generate trajectories.
The \emph{Behavior Reviewer Agent} then enforces instruction alignment and traffic realism through a feedback loop.
After editing the scene graph, a coarse renderer produces an initial video, which is further harmonized by a custom \emph{Video Diffusion Tool} to address lighting inconsistencies and novel-view artifacts.
However, this harmonization may alter the appearance of inserted vehicles. Therefore, a \emph{Video Reviewer Agent} iteratively adjusts diffusion strength and guidance to balance photorealism and appearance preservation.
As shown in Figure~\ref{fig:teaser}, this feedback-driven, multi-modal design achieves photorealism, instruction alignment, structure preservation, and traffic realism simultaneously, significantly outperforming both world models and prior agentic pipelines.

\noindent\textbf{Contributions.} Our main contributions are:

\begin{itemize}[topsep=0pt,itemsep=0pt,parsep=0pt,partopsep=0pt,leftmargin=*]

    \item We introduce LangDriveCTRL, a feedback-driven, natural-language-controllable framework for fine-grained object-level editing of driving videos. It supports object removal, insertion, replacement, and multi-object behavior editing.
    
    \item We design two novel multi-modal reasoning agents: 1) an \emph{Object Grounding Agent} for open-vocabulary object querying, and 2) a \emph{Behavior Editing Agent} for multi-object trajectory generation.
    
    \item 
    We propose feedback-driven iterative refinement of behavior and video via a \emph{Behavior Reviewer Agent} and a \emph{Video Reviewer Agent}, improving both instruction alignment and traffic realism.
    
    \item 
    Extensive experiments demonstrate that LangDriveCTRL achieves nearly 2× higher instruction alignment than prior state-of-the-art methods and significantly improves structural preservation, photorealism, and traffic realism. Meanwhile, it maintains comparable latency to existing SoTA approaches.
    
\end{itemize}

\section{Related Work}

\begin{figure*}[t!]
    \centering
    \includegraphics[width=\linewidth]{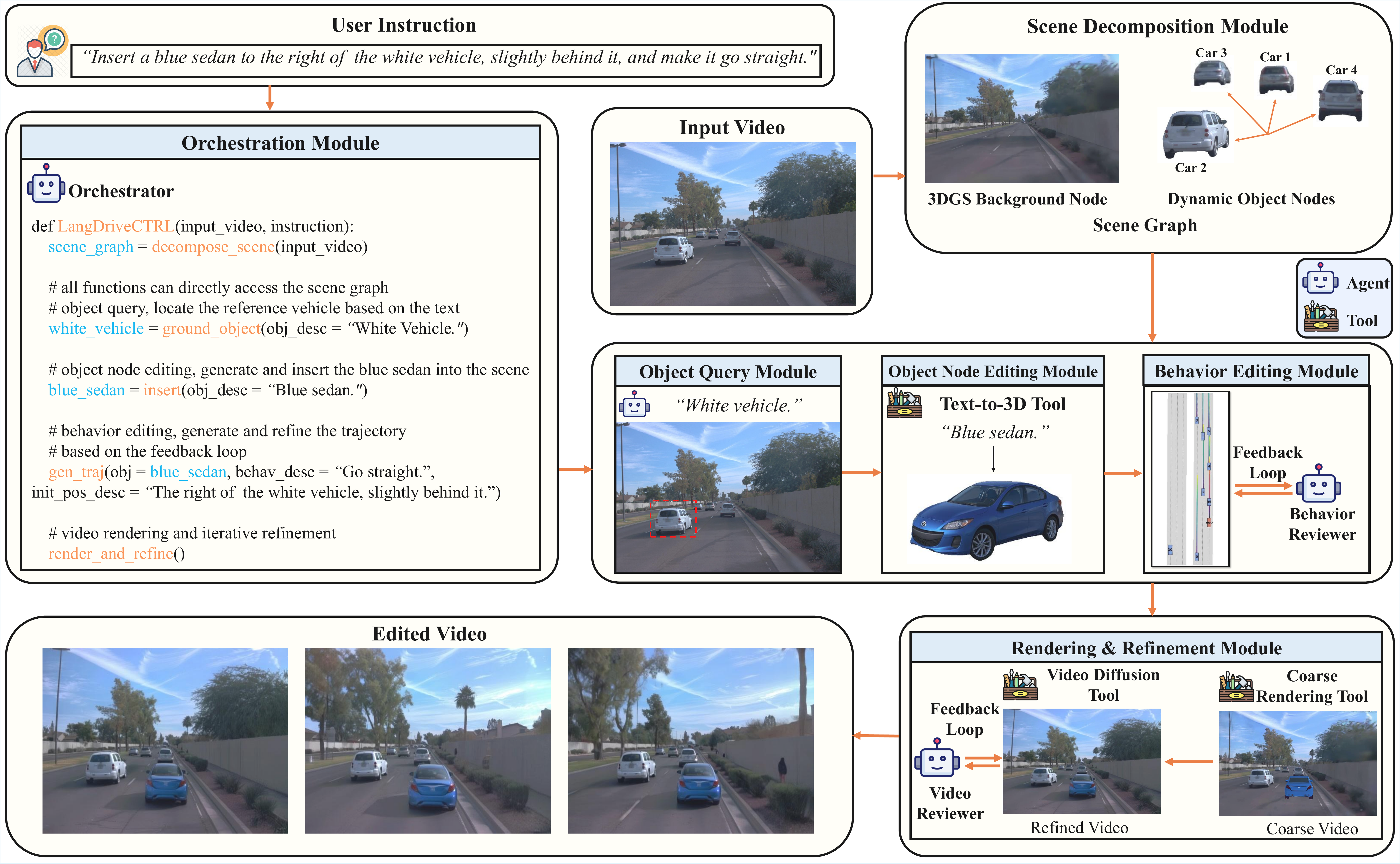}
    \caption{\textbf{Overall Pipeline.} Given an input video and the user instruction, our pipeline first builds a scene graph, which decomposes the scene into a static background node and multiple dynamic object nodes with their trajectories. 
    To execute the instruction, the orchestrator coordinates agents and tools from different modules to work together: the object query module localizes target objects in the scene graph based on textual descriptions; the object node editing module performs object removal, insertion, and replacement; the behavior editing module generates and refines multi-object trajectories based on a feedback loop; finally, the rendering and refinement module renders the edited scene graph and iteratively refines it with a video diffusion tool.
    While the figure illustrates single-object editing, our pipeline is capable of multi-object editing.
    }
    \label{fig:pipeline}
    \vspace{-1.0em}
\end{figure*}

\mypara{Neural Rendering for Driving Scene Editing.}
Neural rendering methods ~\cite{mildenhall2021nerf,tancik2022block,tasneem2024decentnerfs,kerbl20233d,he2022density,he2023grad} such as NeRF and 3D Gaussian Splatting have been widely adopted for autonomous driving due to their ability to reconstruct compositional 3D scenes and support for object-level editing~\cite{chen2024omnire,sun2024lidarf,xiong2025drivinggaussian++}. 
While these optimization-based approaches enable modular manipulation of foreground objects and background, they struggle under significant view changes, lack multi-object simulation capabilities, and do not natively support lighting-aware insertion of new objects.

\vspace{-1.0em}
\mypara{Diffusion Models for Driving Scene Editing.}
Recent editing methods combine neural rendering  with diffusion models \cite{zhang2023adding,song2020denoising,liang2025diffusion} for improved robustness to viewpoint changes and lighting-aware object insertion~\cite{zhu2025scenecrafter,hassan2025gem,liang2025driveeditor,zhao2025drivedreamer}. These pipelines, however, are usually controlled through low-level parameters or 2D/3D bounding boxes, not natural language. 
In contrast, the purely generative world model~\cite{ali2025world} can take natural language instructions and edit videos directly in pixel space, 
but it lacks fine-grained object-level control and often alters the underlying scene structure of the input videos. 

\vspace{-1.0em}
\mypara{Natural-Language-Controllable Simulation.}
LLM-driven modular simulation pipelines ~\cite{wei2024chatdyn,wei2024editable,hsu2025autovfx} leverage LLMs to provide natural-language control over object-level operations (e.g., removal, insertion and replacement). 
However, they struggle with accurate target object localization and realistic trajectory generation due to unimodal text reasoning, produce poor rendering quality from naive compositing, and lack iterative refinement.

\section{Our approach}

Our framework follows an agentic pipeline, as shown in Figure~\ref{fig:pipeline}, that determines which agents (with reasoning ability) and tools (without reasoning ability) to invoke based on user instructions. 
The pipeline consists of different modules to ensure controllability and interpretability while achieving high realism.

\vspace{-1em}
\subsection{Input}

Our pipeline takes a driving video, a user instruction and the scene map as input. 
We assume that the original object trajectories and map are provided.

\vspace{-1em}
\subsection{Orchestration Module}

\mypara{Orchestrator Agent.}
The orchestrator is the central agent in our system that controls the overall workflow. 
It is implemented using an off-the-shelf LLM \cite{achiam2023gpt} that we configure using in-context learning \cite{brown2020language}, and it produces executable Python scripts that call other agents or tools provided in various modules of our system, as shown in Figure~\ref{fig:pipeline}. 

The orchestrator first decomposes the user instruction into sub-instructions for each target object, then designs execution workflows for each object and invokes the corresponding agents and tools from different modules.
To enable this, we encapsulate the operations provided by various modules in our system into modular functions that can be easily assembled into executable scripts. 
We use in-context learning \cite{brown2020language} to teach the LLM how to call these functions and generate executable scripts to fulfill user instructions.

The execution workflow proceeds as follows.
First, the orchestrator employs the scene reconstruction tool to decompose the 3D scene into a static background node and dynamic object nodes with associated trajectories, generating a scene graph.
This scene graph is then shared across all target objects for subsequent editing operations.
For each target object, the orchestrator invokes the object grounding agent to locate the target node in the scene graph based on the textual description. 
Next, depending on the editing type (removal, insertion, or replacement), it calls the appropriate tool or agent to modify the node and update the scene graph.
If the instruction involves trajectory editing, the orchestrator invokes the behavior editing agent to generate trajectories, which are then checked and iteratively refined by a behavior reviewer agent. 
Finally, after all objects have been processed, the orchestrator calls the coarse rendering tool to generate a coarse video, which is then harmonized by the video diffusion tool. 
A video reviewer agent iteratively refines the result to achieve both photorealism and appearance alignment.

\vspace{-1em}
\subsection{Scene Decomposition Module}

The goal of this module is to decompose the input driving video into 
a scene graph $SG$ that enables object-level reasoning and 
controllable editing. 
The scene graph contains a static background node and multiple dynamic object nodes representing vehicles and pedestrians, providing 
a modular and interpretable representation for fine-grained editing.

\mypara{Scene Reconstruction Tool.}
3D Gaussian Splatting (3DGS) \cite{kerbl20233d} is good at representing and rendering static scenes with high photorealism.
Following \cite{chen2024omnire}, the tool decomposes the scene into static background Gaussians and canonical object nodes with trajectory-based transformations. 
These components form a scene graph:
\[
SG(t) = \{ N_\mathrm{bg}, \{ N^\mathrm{i}_\mathrm{asset}(t)\}_{i=1}^{K}\, \},
\]
where $N_\mathrm{bg}$ represents the static background Gaussian primitives, and $N^{\mathrm{i}}_\mathrm{asset}(t)$ are time-dependent object nodes with node ID $\mathrm{i}$.  
Each canonical object node is transformed by its pose that captures its motion trajectory. 
This formulation preserves the spatiotemporal consistency of real-world trajectories while enabling fine-grained, per-object editing.

\vspace{-1em}
\subsection{Object Query Module}
The object query module establishes correspondence between textual object descriptions and scene graph nodes through attribute-based reasoning.
To achieve this goal, previous methods can be roughly categorized into two types: 
open-vocabulary detection/tracking algorithms \cite{ren2024grounded,liu2024grounding,yang2023track,cheng2023segment} and 3DGS-based approaches \cite{qin2024langsplat,li20254d,shi2024language}. 
Although these methods perform well at category-level recognition, they struggle with attribute-based distinctions (e.g., color, type, spatial relationship, and motion). 

\mypara{Object Grounding Agent.} 
To address this limitation, we design an object grounding agent powered by the vision-language model (VLM) \cite{hurst2024gpt}.
It receives three types of information from the input videos, scene graphs and maps to locate target object nodes: 1) appearance information: each node is projected into pixel space and segmented using SAM \cite{kirillov2023segment} to extract its visual appearance; 2) behavior information: motion descriptions are generated from trajectory analysis (speed/heading/lane changes) using heuristic rules (please refer to Section \ref{sec:behavior_description_and_validation} for details); 3)
position information: trajectory coordinates and lane information are used for spatial relationship analysis.
Given all this context information, the agent identifies the target node through a two-stage process. 
First, it decomposes the query into a triplet: reference node, target node, and their spatial relation (e.g.,  ``the black SUV on the left of ego": <ego, black SUV, left>). 
Then, it locates the reference node by matching appearance and behavior, filters candidates by spatial relation, and selects the target node through the same matching process. 
In Table \ref{tab:object_grounding}, we show the superior object grounding performance of this agent w.r.t \cite{ren2024grounded,li20254d}.

\vspace{-1em}
\subsection{Object Node Editing Module}

After identifying the target node, editing operations (e.g., removal/insertion/replacement) can be easily performed by corresponding tools and agents.
\mypara{Removal Tool.}
It removes all Gaussian primitives belonging to the target node, and updates the scene graph.
\mypara{Insertion Agent.}
It first invokes a text-to-3D tool (i.e., Hunyuan3D \cite{zhao2025hunyuan3d}) to generate a mesh, then adjusts its size and local coordinate system to align it with the scene. Finally, the mesh is added to the scene graph as a new node.
For size adjustment, the agent first calculates the mesh's bounding box and rescales it to the scene's actual size. 
For orientation alignment, the agent aligns the mesh's local coordinate system with the scene's world coordinate system by: 1) rendering the mesh from a fixed axis, 2) analyzing its facing direction in the rendered image, 3) determining the local axes.
\mypara{Replacement Agent.}
It essentially combines the removal and insertion operations to replace an existing object node with a new one, while the new node inherits the original trajectory.

\vspace{-1em}
\subsection{Behavior Editing Module}

The behavior editing process involves two specialized agents. The Behavior Editing Agent generates a counterfactual behavior combination list for each object node based on its original trajectory and map information, then selects the behavior combination that best matches the instruction. 
It uses the selected result to generate trajectories through a multi-object simulation tool \cite{chang2025langtraj}. 
The Behavior Reviewer Agent then checks the generated trajectories and performs iterative refinement to ensure instruction alignment and traffic realism.

\mypara{Behavior Editing Agent.}
The agent first uses heuristic tools to generate behavior description of the object's original trajectory. 
This is essentially a behavior combination that describes all matched behaviors (e.g., ``slow down, change from
the middle lane to the left lane, turn left"). 
The next step is \textit{counterfactual behavior generation}. 
Replace/remove/keep/add operations are applied to each behavior in the combination to produce new behavior combinations, which form a combination list. 
These combinations are then filtered using the map to remove unreasonable behaviors, such as ask a vehicle to make a turn when there is no intersection.
Additionally, mutually contradictory behaviors are also filtered out, such as combinations containing both ``going straight" and ``static" (please refer to Section \ref{sec:behavior_description_and_validation} for details).
Finally, the agent selects the best match from the combination list according to the user instruction and original behavior. 
The selected result is then used as the text condition for LangTraj \cite{chang2025langtraj}, a diffusion-based, language-conditioned trajectory simulator for multi-object simulation.
Importantly, the selection is a behavior combination rather than a single behavior (e.g., if the user instruction is ``speed up", and the original behavior is ``slow down, change from
the middle lane to the left lane, turn left", the selected behavior combination will be ``speed up, change from
the middle lane to the left lane, turn left"). 
The purpose of doing this is twofold: 1) to filter out unreasonable behaviors; 2) to preserve the object's original behaviors as much as possible. For example, if the original behavior is ``go straight" and the user asks the vehicle to slow down, the vehicle should maintain going straight while slowing down.
Please refer to Table \ref{tab:ablation_cf_behavior} for an ablation study on the counterfactual behavior generation component.

\mypara{Behavior Reviewer Agent.}
However, generated trajectories from LangTraj \cite{chang2025langtraj} may not align with instructions and may involve collisions or off-road scenarios. 
The reviewer agent addresses this through an automatic feedback loop that iteratively validates and refines trajectories. Specifically, it employs trajectory validation functions to evaluate the instruction alignment and traffic rule compliance (please refer to Section \ref{sec:behavior_description_and_validation} for the details of validation functions). 
For multi-object simulation, the agent handles successful and unsuccessful objects differently.
For objects that already satisfy all requirements, it stores their successful trajectories and uses them as guidance for subsequent generations. 
This makes it easier to achieve trajectory-instruction alignment and also enables interaction with other objects.
For objects that do not meet the requirements, the agent adjusts the guidance configuration for LangTraj \cite{chang2025langtraj} accordingly.
Specifically, if behavior misaligns with the instruction, it increases the classifier-free guidance weight. 
For off-road or collision violations, it adds the corresponding off-road or collision avoidance guidance and adjusts its weight to improve traffic compliance.
Based on this feedback loop, the module can consistently generate realistic and accurate trajectories.

\begin{figure*}[t]
    \vspace{-0.2cm}
    \centering
    \includegraphics[width=\textwidth]{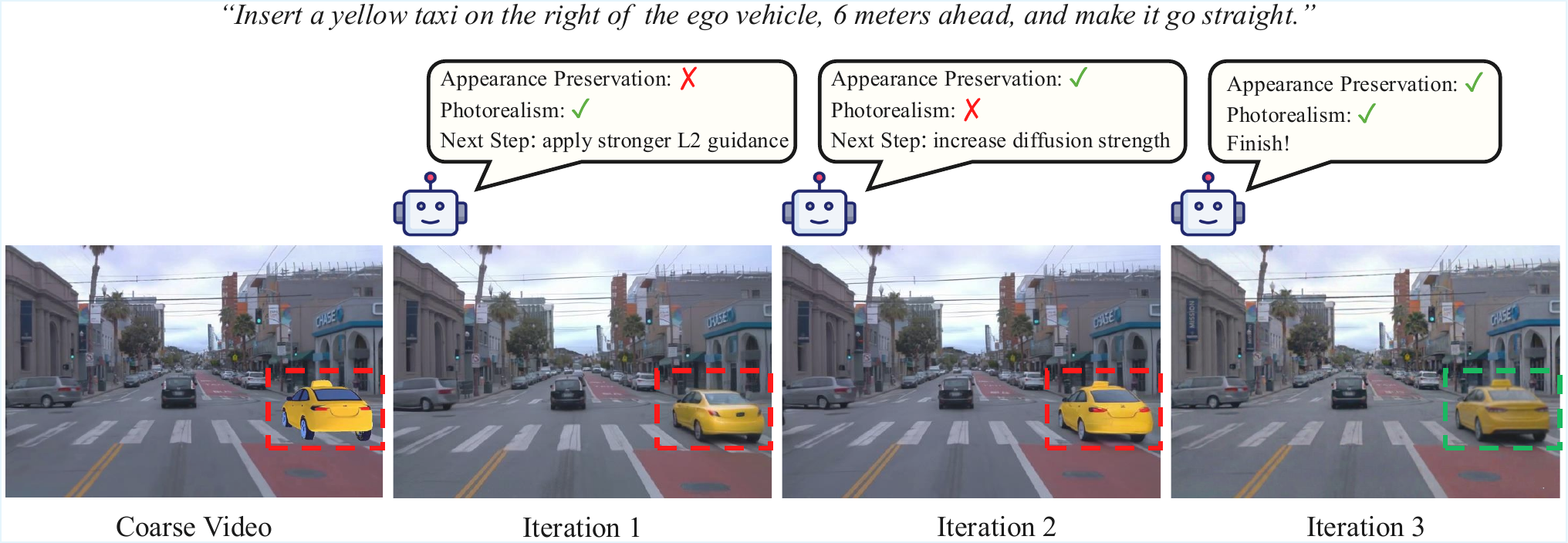}
    \caption{\textbf{Effect of the video reviewer agent.} The video reviewer dynamically adjusts the diffusion strength and the L2 guidance weight based on feedback. \textit{Iteration 1:} The taxi looks realistic, but appearance preservation is compromised (the roof light disappears), so the agent increases the L2 guidance weight. \textit{Iteration 2:} The appearance is preserved, but it looks unrealistic (too bright), so the agent increases the diffusion strength. \textit{Iteration 3:} Both photorealism and appearance preservation are achieved, and the agent stops the refinement.}
    \label{fig:video_iterative_refinement}
    \vspace{-1.0em}
\end{figure*}

\vspace{-1em}
\subsection{Rendering and Refinement Module}

\mypara{Coarse Rendering Tool.}
This tool renders the edited scene from the updated scene graph.
Specifically, it renders the 3DGS based scene graph using the rasterization algorithm \cite{kerbl20233d}, and renders the inserted object meshes using PyVista \cite{sullivan2019pyvista}. 
The rendered components are then composited with depth information.
However, videos generated by this tool typically lack photorealism. 
The newly inserted objects often appear unnatural, and when new viewpoints differ significantly from the original ones (e.g., when modifying the ego vehicle's trajectory), the rendering quality of 3DGS drops quickly.

\mypara{Video Diffusion Tool.}
To address the quality issue, this tool employs a video diffusion model that takes the coarse video as condition to generate the enhanced output.
Specifically, it adopts CogVideoX \cite{yang2024cogvideox} as the backbone and finetunes the model using two strategies: 1) replacing Gaussian primitives in the 3DGS representation with object meshes to learn the photorealistic vehicle appearances, 2) training on noisy Gaussian rendering pairs curated via cycle reconstruction strategy~\cite{wu2025difix3d+} for effective denoising.

\mypara{Video Reviewer Agent.}
However, while the video diffusion tool can generate photorealistic results, it may alter the appearance (shape, key parts, type, color, etc) of inserted vehicles. 
In diffusion processes, higher denoising strength (i.e., greater noise levels) generally produces 
more realistic outputs but risks losing information from the conditioning input \cite{brooks2023instructpix2pix,meng2021sdedit}. 
For example, in Iteration 1 of Figure~\ref{fig:video_iterative_refinement}, the yellow taxi's roof light disappears. 
To improve video quality while preserving vehicle 
appearance, this VLM powered agent employs feedback-driven iterative refinement. 
It dynamically adjusts the denoising strength to control photorealism and tunes the L2 guidance loss weight to preserve object appearance.
The L2 guidance loss is computed at each denoising step by measuring the L2 distance (in latent space) between the inserted-vehicle regions of the predicted and the condition video.

The iterative refinement process works as follows. First, the agent applies diffusion with relatively high denoising strength, which empirically produces photorealistic results. 
The agent then reviews the output. 
If appearance is compromised (e.g. key parts are missing, shape/color changes), it increases the L2 guidance weight.
Conversely, if the inserted vehicle appears unrealistic (e.g., lighting mismatches the environment), it increases the denoising strength. 
This process continues until both photorealism and appearance preservation are satisfied, or the maximum iteration number is reached, as shown in Figure~\ref{fig:video_iterative_refinement}.

\section{Experiments}
\label{sec:exp}
In this section, we provide quantitative and qualitative comparisons with baselines. 
For better visualization, please refer to the videos in the project page.

\begin{figure*}[t!]
    \centering
    \includegraphics[width=\textwidth]{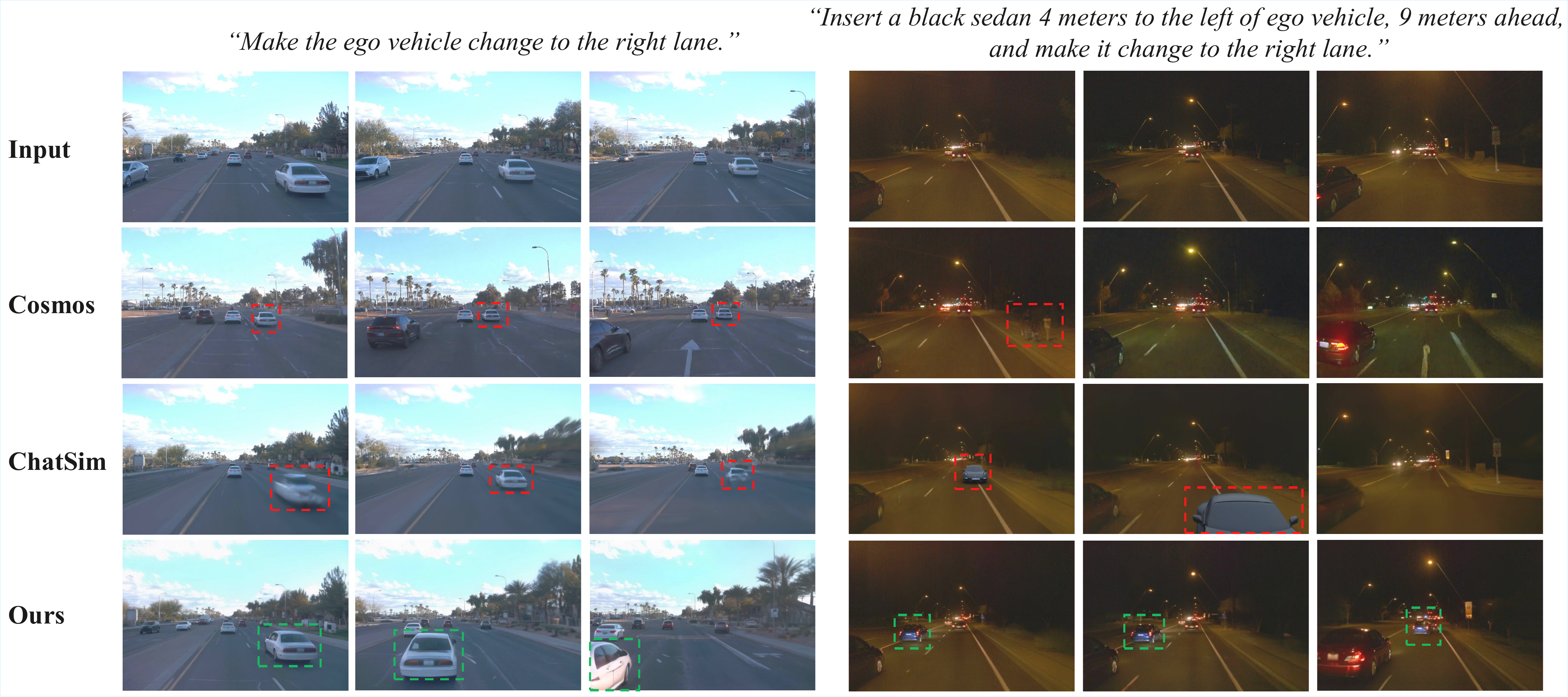}
    \caption{\textbf{Qualitative comparison with baselines}. The results generated by Cosmos \cite{ali2025world} fail to align with the instruction and do not preserve the background well. 
    ChatSim \cite{wei2024editable} produces editing results with poor visual quality, inaccurate trajectories, and collision issues. 
    Our method clearly outperforms them in photorealism, instruction alignment, structure preservation, and traffic realism.}
    \label{fig:base_exp}
    \vspace{-1.5em}
\end{figure*}

\vspace{-1em}
\subsection{Evaluation Metrics}

We evaluate our method on the following five aspects.
1) \textbf{Photorealism.} We use \textit{FID} \cite{heusel2017gans} to assess image realism, and \textit{FVD} \cite{unterthiner2018towards} to evaluate temporal consistency and overall video quality.
2) \textbf{Instruction Alignment.}  
We measure instruction alignment using the following two metrics. 
For \textit{Appearance Alignment} metric, we sample frames from both the original and edited videos. 
We then use the GPT-4o \cite{hurst2024gpt} to compare them and determine whether the target object has been accurately deleted, inserted, or replaced.
For \textit{Behavior alignment} metric, we first use the Grounded-SAM-2 \cite{ren2024grounded} model to track the edited object, and then back-project its trajectory from pixel coordinates to world coordinates. 
Finally, based on the map information, we evaluate whether the trajectory matches the instructions.
3) \textbf{Structure Preservation.}  
Following \cite{li2025five}, we use the self-similarity matrix from DINO  \cite{caron2021emerging} to capture the structural information of images. 
We then compare the difference between the matrices of the original and edited images.
4) \textbf{Traffic Realism.}  
The generated trajectories should not violate traffic rules.
Therefore, we also report the \textit{Collision Rate} and \textit{Off-road Rate} by sampling frames from edited videos and using the GPT-4o \cite{hurst2024gpt} to detect such incidents.
5) \textbf{User Study.}  
For all four aspects mentioned above, we also conduct human evaluation with 26 participants. 
For each aspect, participants are asked to select which method performs best among the three methods and ``none''.

\vspace{-1em}
\subsection{Experiment Settings}

\mypara{Dataset and Instructions.}
We curate 30 diverse scenes from a real-world driving dataset (Waymo Open Dataset \cite{sun2020scalability}), covering different times of day, road types, and weather conditions. 
Our work primarily focuses on vehicle editing in driving scenes, so we select scenes with fewer pedestrians.
Detailed scene IDs are provided in Table \ref{tab:test_scenes}.
For test instructions, we generate them using GPT-4 \cite{achiam2023gpt}, followed by human filtering. 
For each scene, we generate 4 types of instructions (with 2-3 instructions per type): 1) \textit{removal} (e.g., ``remove the blue sedan in front''); 2) \textit{replacement}  (e.g., ``replace the van on the right with a yellow taxi''); 3) \textit{behavior editing} (e.g., ``make the ego vehicle turn left''); 4) \textit{insertion} (e.g., ``insert a black SUV 10 meters in front of the ego vehicle and make it change to the right lane'').

\begin{table*}[t]
  \centering
  \caption{\textbf{Quantitative comparison with baselines.} Our method consistently outperforms all baselines across all metrics while maintaining efficiency. Abbreviations: App. = Appearance, Beh. = Behavior, Str. = Structure Distance, Col. = Collision, Off. = Off-road. User (\%) shows the percentage of user study participants who choose each method as the best for that aspect, with options including the three methods and "none". Editing time is measured in minutes per scene on a single A6000 GPU.}
  \label{tab:base_exp}
  \vspace{-1em}
  \setlength{\tabcolsep}{2.5pt}   
  \small                    
  \resizebox{\textwidth}{!}{%
  \begin{tabular}{l
                  lll      
                  lll      
                  ll       
                  lll      
                  l}       
    \toprule
    & \multicolumn{3}{c}{Photorealism}
    & \multicolumn{3}{c}{Instr. Align.}
    & \multicolumn{2}{c}{Struct. Pres.}
    & \multicolumn{3}{c}{Traffic Realism}
    & Efficiency \\
    \cmidrule(r){2-4}\cmidrule(r){5-7}\cmidrule(r){8-9}\cmidrule(r){10-12}\cmidrule(r){13-13}
    Method
      & FID $\downarrow$ & FVD $\downarrow$ & User (\%) $\uparrow$
      & App. (\%) $\uparrow$ & Beh. (\%) $\uparrow$ & User (\%) $\uparrow$
      & Str. $\downarrow$ & User (\%) $\uparrow$
      & Col. (\%) $\downarrow$ & Off. (\%) $\downarrow$ & User (\%) $\uparrow$
      & Time (min) $\downarrow$ \\
    \midrule
    Cosmos \cite{ali2025world}   
      & 33.42 & 797.51 & 34.60
      & 46.25 & 32.86 & 10.50
      & 74.07 & 19.10
      & 3.16 & 3.95 & 35.5
      & \textbf{16.9} \\ 
    ChatSim \cite{wei2024editable}  
      & 47.70 & 605.69 & 4.80
      & 42.33 & 26.64 & 3.90
      & 46.52 & 7.40
      & 27.62 & 24.71 & 5.60
      & 19.3 \\
    \textbf{Ours} 
      & \textbf{32.85} & \textbf{467.20} & \textbf{54.20}
      & \textbf{82.19} & \textbf{71.67} & \textbf{60.40}
      & \textbf{34.62} & \textbf{65.00}
      & \textbf{0.58} & \textbf{1.73} & \textbf{48.30}
      & 17.1 \\
    \bottomrule
  \end{tabular}}
  \vspace{-1.5em}
\end{table*}

\mypara{Baselines.}
We use both LLM agent-based (ChatSim \cite{wei2024editable}) and diffusion model-based (Cosmos \cite{ali2025world}) driving scene editors as baselines.
Both are state-of-the-art open-source methods in their respective categories.
To ensure fair comparison, we strictly follow the original settings of each method. 
We evaluate all methods on the same scene and instruction pairs.
In addition to existing driving scene editing methods, we also construct a baseline by naively combining an image editing method with an image-to-video method. 
Specifically, we first use ChronoEdit \cite{wu2025chronoedit} to edit the first frame, then apply Wan 2.2 \cite{wan2025wan} to convert the edited first frame into a video, with both stages conditioned on the instruction. 
Please refer to Section \ref{sec:chronoedit_wan} for details.

\mypara{Implementation Details.}

We use the front camera of scenes for experiments. 
The input videos are 8 seconds long with an FPS of 10. 
For the LLM model, we use GPT-4 \cite{achiam2023gpt}, and for the VLM model, we use GPT-4o \cite{hurst2024gpt}. 
For the behavior and video reviewer agents, we set the maximum iteration number for the feedback loop as 5.
For the video diffusion tool, we adopt CogVideoX~\cite{yang2024cogvideox} as the backbone and initialize the model with pretrained weights from TrajectoryCrafter~\cite{yu2025trajectorycrafter}. 
We further fine-tune it in two stages: 1) 40k iterations on 33-frame short videos, followed by 2) 20k iterations on 81-frame long videos. 
Both stages are trained with a batch size of 4 on a 4$\times$H100 GPU workstation, for 48 hours each.

\vspace{-1em}
\subsection{Quantitative Comparison with Baselines}
\mypara{Editing Performance.} We report editing performance metrics across four key aspects in Table~\ref{tab:base_exp}.
As can be seen, our method outperforms the baselines across all metrics, particularly in appearance and behavior alignment. 
In terms of photorealism and structure preservation, our editing results not only achieve the highest visual quality and temporal consistency, but also preserve the original structure well. 
While Cosmos \cite{ali2025world} demonstrates good photorealism on individual frames, it tends to modify the background simultaneously. 
ChatSim \cite{wei2024editable}, on the other hand, shows poor performance in both visual quality and structure preservation. 
For instruction alignment,
both Cosmos \cite{ali2025world} and ChatSim \cite{wei2024editable} fail to accurately remove, insert, or replace objects, and cannot generate precise trajectories for target behaviors.
In contrast, our method substantially outperforms them in both appearance and behavior alignment. 
Regarding traffic realism, our method effectively models multi-object interactions, thus significantly reducing traffic violations such as collisions and off-road incidents. 
For user study, our method also greatly outperforms baselines, particularly in instruction alignment and structure preservation.

\begin{figure}[t]
  \vspace{-0.2cm}
  \centering
  \includegraphics[width=\columnwidth]{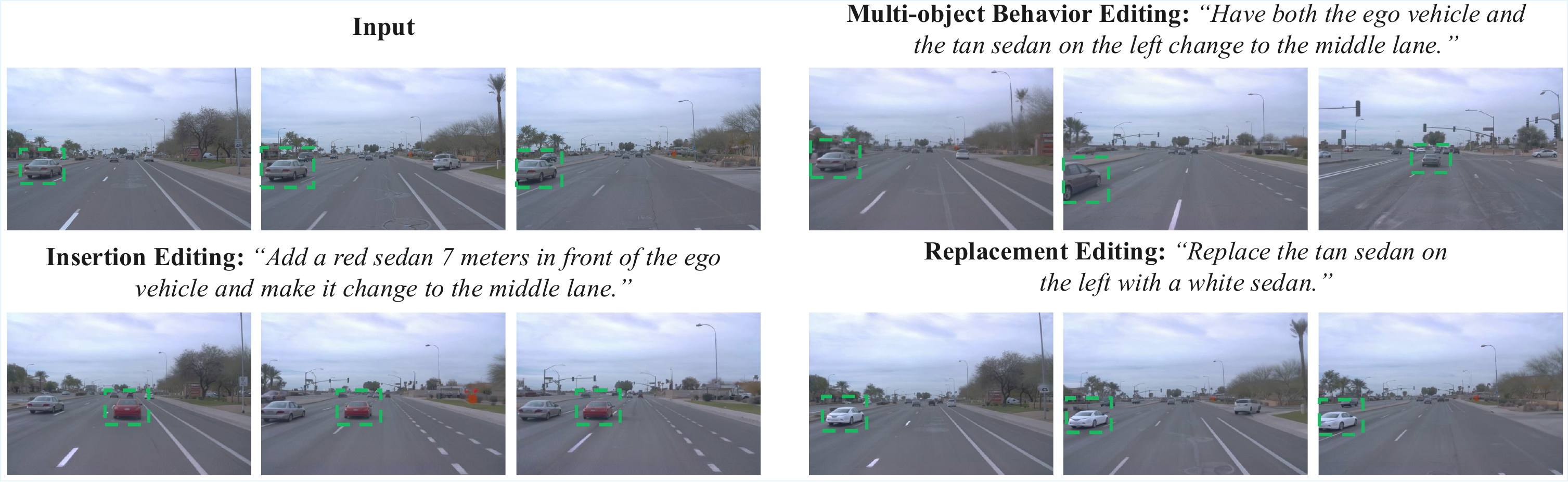}
  \vspace{-2em}
  \caption{\textbf{Qualitative editing results.} We demonstrate our method's editing capabilities for diverse scenario generation.
  Note that for better visualization, the timestamps are not strictly aligned.}
  \label{fig:editing_examples}
  \vspace{-2em}
\end{figure}

\mypara{Computational Efficiency.}
We also report editing time for each method in Table~\ref{tab:base_exp}. All methods generate 8-second videos at 10 fps 
on a single NVIDIA A6000 GPU. 
Our method requires 17.1 minutes per edit, 
comparable to baselines (Cosmos: 16.9 minutes, ChatSim: 19.3 minutes). 
Note that both ChatSim and our method require a one-time 3D reconstruction preprocessing step per scene, which takes around 2 hours.
We further analyze per-module timing for our method in Table~\ref{tab:timing}. 
The object node editing module (dominated by Text-to-3D Tool) and video iterative refinement are the most time-consuming components.

\vspace{-1.5em}
\begin{table}[ht]
  \centering
  \caption{\textbf{Per-module timing for our method.} Object node editing and video iterative refinement dominate the runtime. 
  Note that behavior editing time already includes the feedback loop. All timings measured on a single A6000 GPU.}
  \vspace{-0.5em}
  \label{tab:timing}
  \setlength{\tabcolsep}{2pt}
  \footnotesize
  \resizebox{\columnwidth}{!}{%
  \begin{tabular}{lccccc}
    \toprule
    Ours
      & Object Query
      & Object Node Editing
      & Behavior Editing
      & Coarse Rendering
      & Video Iterative Refinement \\
    \midrule
    Time
      & 1.4 mins
      & 6.1 mins
      & 1.5 mins
      & 0.5 mins
      & 7.6 mins \\
    \bottomrule
  \end{tabular}}
  \vspace{-3em}
\end{table}

 \vspace{-1em}
\subsection{Qualitative Comparison with Baselines}
We provide visual comparison results in Figure \ref{fig:base_exp}. 
The first example involves ego vehicle trajectory editing (``Make the ego vehicle change to the rightmost lane."). 
We not only achieve accurate ego view changes, but also capture the surrounding environmental lighting information (e.g., realistic highlights and reflections on the vehicle body). 
In contrast, both Cosmos \cite{ali2025world} and ChatSim \cite{wei2024editable} fail to perform accurate ego vehicle trajectory editing. 
ChatSim \cite{wei2024editable} also suffers from poor visual quality (e.g., artifacts in moving objects). 

In the second example (``Insert a black sedan 4 meters to the left of ego vehicle, 9 meters ahead, and make it change to the right lane."), we use a more challenging scenario at nighttime. 
Our editing results show that the inserted vehicle not only executes the lane change accurately, but also seamlessly adapts to the nighttime lighting. Cosmos \cite{ali2025world} not only fails to insert the requested vehicle but also adds unrequested pedestrians, completely disregarding the instruction. 
Although ChatSim \cite{wei2024editable} correctly inserts a black sedan, the result looks highly unnatural. 
Moreover, the generated trajectory does not follow the target behavior, and the inserted vehicle collides with existing vehicles, failing to achieve proper multi-vehicle interaction.

\vspace{-1em}
\subsection{Diverse Editing Capabilities}

In Figure~\ref{fig:editing_examples}, we show our method’s diverse editing capabilities, including multi-object behavior editing, object-level insertion and replacement. 
The examples illustrate that the object grounding agent reliably localizes the target object nodes based on textual descriptions (``tan sedan on the left''), while the behavior editing module correctly modifies the trajectories of both the ego vehicle and the referenced sedan according to the instructions.

\vspace{-1em}
\subsection{Ablation Studies}

\begin{figure*}[t]
    \centering
    \includegraphics[width=\textwidth]{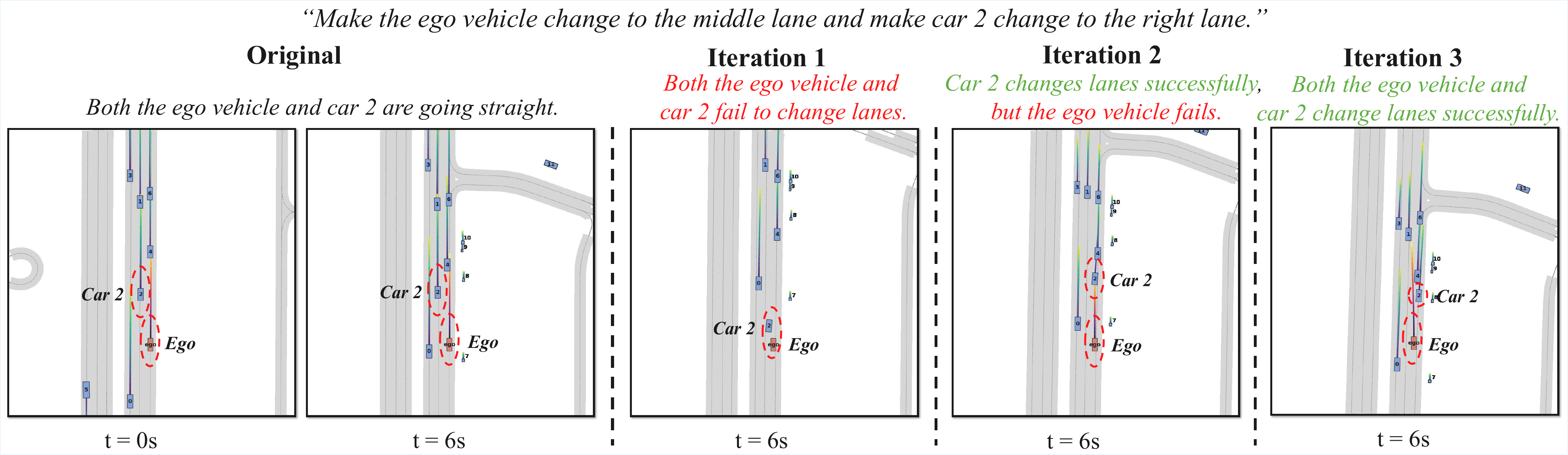}
    \vspace{-2em}
    \caption{\textbf{Effect of the behavior iterative refinement.} The feedback loop effectively improves the alignment between generated trajectories and instructions while avoiding off-road behavior and collisions.}
    \label{fig:behavior_feedback_loop}
    \vspace{-2em}
\end{figure*}

To validate the effectiveness of our behavior and video refinement modules (the behavior reviewer agent and the video reviewer agent), we conduct ablation studies on these two components. 
For behavior refinement experiments, we use the same 30 test scenes as in Table \ref{tab:base_exp}, but generate new test instructions (70 in total). 
While the behavior editing instructions from Table \ref{tab:base_exp} mostly target single objects, here we create more challenging instructions that specify target behaviors for multiple objects simultaneously. 
Additionally, during evaluation, we directly evaluate the generated trajectories instead of the edited videos. 

\begin{table}[htbp]
  \centering
  \vspace{-1.5em}
  \caption{\textbf{Ablation study for iterative behavior refinement.} Overall success indicates that behavior alignment is achieved while avoiding both collisions and off-road behavior. With iterative behavior refinement, the overall success rate of trajectory generation increases significantly.}
  \vspace{-0.5em}
  \label{tab:behavior_feedback_loop}
  \resizebox{0.97\columnwidth}{!}{%
  \begin{tabular}{lcccc}
    \toprule
    Behavior Refinement
    & Behavior Alignment (\%) $\uparrow$
    & Collision (\%) $\downarrow$
    & Off-road (\%) $\downarrow$
    & Overall Success (\%) $\uparrow$ \\
    \midrule
    \xmark & 54.29 & 35.71 & 30.00 & 34.29 \\
    \textbf{\cmark} & \textbf{70.00} & \textbf{22.86} & \textbf{14.29} & \textbf{51.43} \\
    \bottomrule
  \end{tabular}}
  \vspace{-2.0em}
\end{table}

As shown in Table \ref{tab:behavior_feedback_loop}, our feedback loop significantly improves trajectory–inst
ruction alignment and decreases both off-road and collision cases. 
We also present a visual comparison in Figure \ref{fig:behavior_feedback_loop} (``Make the ego vehicle change to the middle lane and make car 2 change to the right lane."). 
In iteration 1, the reviewer checks the generated trajectories and finds that neither the ego vehicle nor car 2 produces target trajectories, so it increases the Classifier-Free Guidance (CFG) weight for both. 
In iteration 2, the reviewer observes that car 2 now generates the correct trajectory, while the ego vehicle still does not. Therefore, it saves car 2's trajectory as guidance for the next iteration and continues to increase the CFG weight for the ego vehicle. 
After iteration 3, both ego vehicle and car 2 generate correct trajectories, so the process stops.

\begin{table}[htbp]
  \vspace{-1.5em}
  \centering
  \caption{\textbf{Ablation study for iterative video refinement.} With iterative video refinement, both photorealism and appearance alignment can be achieved.}
  \vspace{-0.5em}
  \label{tab:vdm}
  \resizebox{0.9\columnwidth}{!}{%
  \begin{tabular}{cccccc}
    \toprule
    \makecell{Video Diffusion Tool} 
    & \makecell{Video Reviewer Agent}
    & FID $\downarrow$ 
    & FVD $\downarrow$ 
    & \makecell{Appearance Alignment (\%) $\uparrow$} \\
    \midrule
    \xmark & \xmark & 43.54 & 613.72 & \textbf{87.69} \\
    \cmark & \xmark & 36.83 & 501.84 & 65.47 \\
    \textbf{\cmark} & \textbf{\cmark} & \textbf{36.78} & \textbf{493.25} & 85.28 \\
    \bottomrule
  \end{tabular}}
  \vspace{-2.0em}
\end{table}

Table~\ref{tab:vdm} provides ablation results for the video iterative refinement. 
To illustrate how video diffusion models may alter the original appearance of inserted vehicles, we use only insertion and replacement instructions in this experiment. 
The coarse video (row 1) aligns well with instructions but exhibits poor photorealism. 
Applying the video diffusion tool once (row 2) for harmonization significantly improves photorealism but compromises appearance alignment. 
Our iterative refinement (row 3) achieves both objectives simultaneously.

\vspace{-1em}
\subsection{Hallucinations of Video Diffusion Tool}
While video diffusion models can improve realism, they may also introduce hallucinations \cite{aithal2024understanding}. 
To analyze potential hallucinations from our video diffusion tool, we evaluate two additional metrics. 
1) We use the 3D Consistency metric from WorldScore \cite{duan2025worldscore}, which measures geometric consistency via depth-based reprojection error across consecutive frames. 
2) We assess road structure preservation using NTL-IoU metric from DriveDreamer4D \cite{zhao2025drivedreamer4d}, which computes the mean IoU between predicted 2D lanes and projected ground-truth 3D lanes. 
We report both metrics before and after video diffusion refinement (i.e., our coarse video vs. refined video), as well as comparisons against all baselines. 
As shown in Table \ref{tab:hallucination_analysis}, video diffusion can introduce mild hallucinations, e.g., slightly degrading geometric consistency and lane structure. 
However, these effects are limited, and our method still outperforms all baselines.

\begin{table}[htbp]
  \centering
  \vspace{-1.5em}
  \caption{\textbf{Hallucination analysis before and after video diffusion refinement.}}
  \vspace{-0.5em}
  \label{tab:hallucination_analysis}
  \resizebox{0.7\columnwidth}{!}{%
  \begin{tabular}{lcccc}
    \toprule
    Metric & Ours (Coarse) & Ours (Refined) & ChatSim \cite{wei2024editable} & Cosmos \cite{ali2025world} \\
    \midrule
    3D Consistency \cite{duan2025worldscore} $\uparrow$ & 74.07 & 72.64 & 71.32 & 69.85 \\
    NTL-IoU \cite{zhao2025drivedreamer4d} $\uparrow$ & 52.11 & 50.96 & 50.13 & 48.76 \\
    \bottomrule
  \end{tabular}}
  \vspace{-2.0em}
\end{table}

\vspace{-1em}
\section{Conclusion}
We introduce LangDriveCTRL, a natural-language-controllable framework for editing real-world driving videos, supporting both object-level operations (removal, insertion, and replacement) and multi-object behavior editing.
We demonstrate through extensive quantitative and qualitative evaluations that our method simultaneously achieves photorealism, instruction alignment, structure preservation, and traffic realism, significantly outperforming prior methods.

\clearpage  

%
%
\bibliographystyle{splncs04}
\bibliography{main}

\clearpage

In the supplementary material, we provide additional experiments, detailed comparison with related work, implementation details, extra qualitative results and failure case.

\section{Additional Experiments}

In this section, we conduct six additional experiments: 1)  performing an ablation study on the \textit{Object Grounding Agent} to verify its contribution; 2) performing an ablation study on the \textit{Behavior Editing Agent} to validate its effectiveness, especially the counterfactual behavior generation component; 3) comparing ``Ours (Coarse)'' against ``ChatSim'' \cite{wei2024editable} by removing the video diffusion tool and video refinement agent, demonstrating that our architecture itself is a significant contribution; 4) constructing an additional baseline by naively combining an image editing method \cite{wu2025chronoedit} with an image-to-video method \cite{wan2025wan}, and comparing against it to further validate the effectiveness of our approach; 5) evaluating our method on the downstream task of object detection; 6) conducting a cross-model judge agreement validation to examine whether different judge models produce consistent preferences.

\vspace{-1em}
\subsection{Ablation on Object Grounding Agent}
\label{sec:ablation_grounding_agent}
For the open-vocabulary object query experiment, we use Grounding SAM \cite{ren2024grounded} and 4DLangSplat \cite{li20254d} as baselines. 
Grounding SAM \cite{ren2024grounded} first performs open-vocabulary detection on images through Grounding DINO \cite{liu2024grounding} to obtain object bounding boxes. 
It then uses SAM \cite{kirillov2023segment} to generate object masks based on these bounding boxes. 
4DLangSplat \cite{li20254d} first reconstructs the dynamic scene through 4D Gaussian Splatting \cite{wu20244d}. Each Gaussian primitive is then augmented with CLIP features \cite{radford2021learning} and caption embeddings
to learn semantic attributes.

To construct the test data, we select 5 scenes from the original 30 test scenes, which cover different times of day, weather conditions, and road types. 
For each scene, we randomly select 10 images and generate one query per image. 
This results in a total of 50 queries. 
To obtain ground truth object masks, we use SAM \cite{kirillov2023segment} for manual annotation. 
Finally, we calculate the IoU between predicted masks and the ground truth masks.
And we consider predictions with IoU greater than 0.2 as successful detections.

\vspace{-1em}
\begin{table}[htbp]
  \centering
  \caption{\textbf{Comparison of object grounding performance across different methods.} Predictions with IoU $>$ 0.2 are considered successful detections. Our method significantly outperforms the baselines.}
  \label{tab:object_grounding}
  \setlength{\tabcolsep}{3pt}
  \scriptsize
  \begin{tabular*}{0.55\columnwidth}{@{\extracolsep{\fill}}lc@{}}
    \toprule
    Method
    & \makecell{Accuracy (\%) $\uparrow$} \\
    \midrule
    Grounding SAM \cite{ren2024grounded} & 44.00 \\
    4DLangSplat \cite{li20254d} & 38.00 \\
    \textbf{Ours} & \textbf{84.00} \\
    \bottomrule
  \end{tabular*}
\end{table}
\vspace{-1em}

\begin{figure*}[t]
    \centering
    \includegraphics[width=\textwidth]{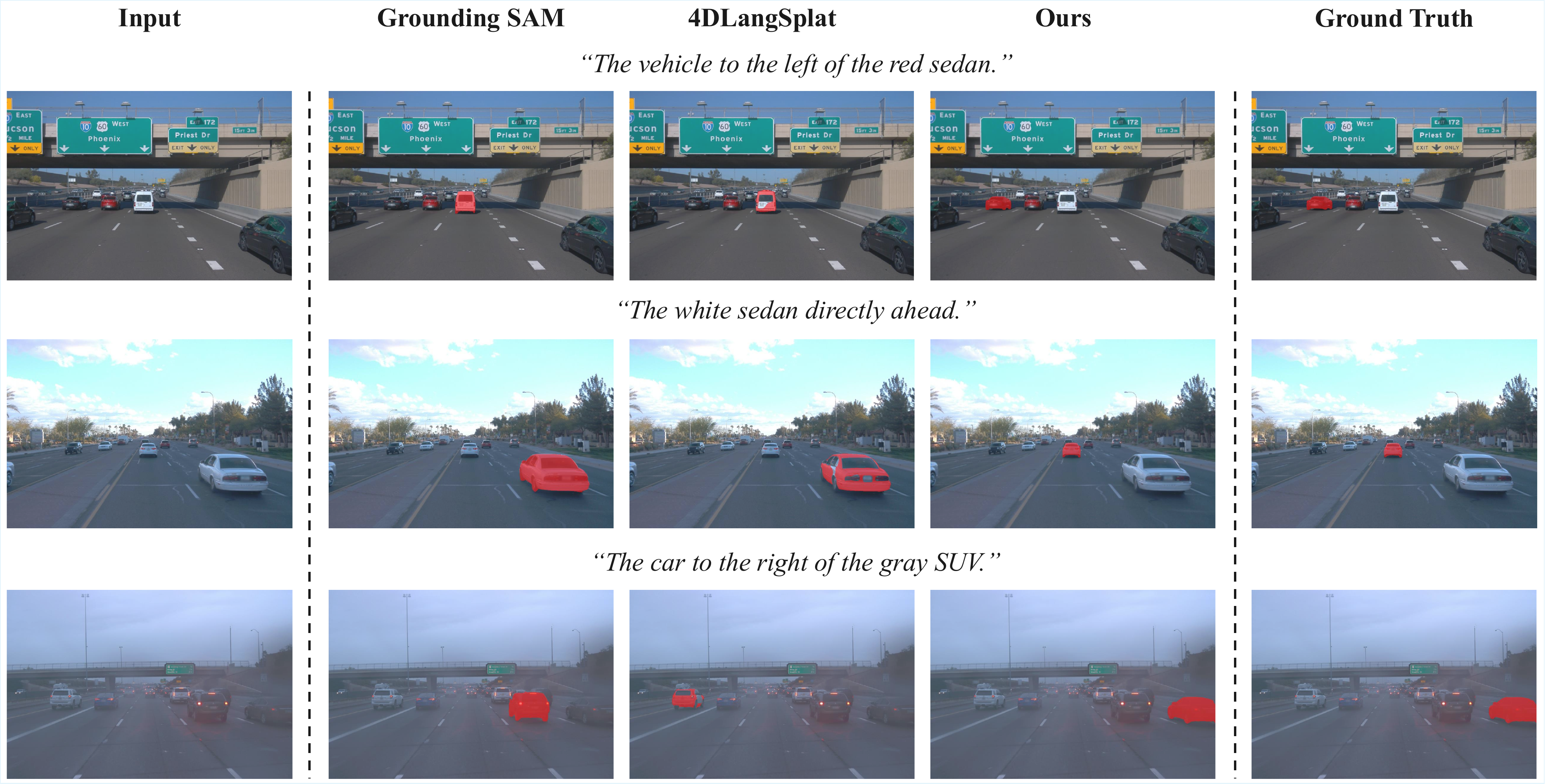}
    \caption{\textbf{Qualitative results for open-vocabulary object query.}
    The detected object masks are highlighted in red.
    Compared to Grounding SAM \cite{ren2024grounded} and 4DLangSplat \cite{li20254d}, our method demonstrates stronger capability in recognizing different attributes of vehicles, especially spatial information.}
    \label{fig:supp_object_query}
    \vspace{-1em}
\end{figure*}

Table~\ref{tab:object_grounding} and Figure~\ref{fig:supp_object_query} present the quantitative and qualitative results of different object grounding methods, respectively. 
As shown, Grounding SAM \cite{ren2024grounded} and 4DLangSplat \cite{li20254d} fail to accurately recognize different object attributes, while our method can do correct reasoning based on appearance, behavior, and position context information.

\vspace{-1em}

\subsection{Ablation on Behavior Editing Agent}
\label{sec:ablation_counterfactual}
We conduct an ablation study on the counterfactual behavior generation component (using the same test instructions as in Table \ref{tab:base_exp} of the main paper). 
When it is removed, the behavior editing agent directly uses the original user instruction as the text condition for the trajectory generator \cite{chang2025langtraj}, without performing any filtering or augmentation of the target behavior.
Table~\ref{tab:ablation_cf_behavior} presents the comparison results. 
With counterfactual behavior generation, our method achieves better behavior alignment while avoiding traffic violations. 
This is because we augment the target behavior with the original behavior description, and since the original trajectory is realistic, it provides prior knowledge to the trajectory generator.

\vspace{-1.0em}
\begin{table}[htbp]
  \centering
  \caption{\textbf{Ablation study on counterfactual behavior generation.} CF Gen.: Counterfactual Generation, Behav. Align.: Behavior Alignment. Overall Success indicates that all requirements (behavior alignment, no collision, on-road) are satisfied. With counterfactual behavior generation, the overall success rate increases significantly.}
  \vspace{-0.5em}
  \label{tab:ablation_cf_behavior}
  \resizebox{0.85\columnwidth}{!}{%
  \begin{tabular}{lcccc}
    \toprule
    CF Gen.
    & Behav. Align. (\%) $\uparrow$
    & Collision (\%) $\downarrow$
    & Off-road (\%) $\downarrow$
    & Overall Success (\%) $\uparrow$ \\
    \midrule
    \xmark & 65.71 & 28.57 & 22.86 & 47.14 \\
    \textbf{\cmark} & \textbf{70.00} & \textbf{22.86} & \textbf{14.29} & \textbf{51.43} \\
    \bottomrule
  \end{tabular}}
  \vspace{-1.0em}
\end{table}

In Table~\ref{tab:ablation_cf_behavior}, all test instructions are reasonable. However, counterfactual behavior generation also plays a crucial role in filtering out unreasonable behaviors. 
We therefore construct 30 unreasonable instructions to test this capability, such as asking vehicles to turn where no intersection exists, or making vehicles in the leftmost lane to change lanes further left. 
The results in Table~\ref{tab:ablation_unreasonable} reveal that without counterfactual behavior generation, the trajectory generator naively accepts unfeasible behaviors and generates trajectories. With counterfactual behavior generation, the vast majority of unreasonable behaviors are filtered out. 
Nevertheless, some failure cases persist. For example, when a vehicle is in the rightmost lane with a median-separated lane on its right, the system may allow it to cross the median barrier.

\vspace{-1.0em}
\begin{table}[h]
\centering
\footnotesize
\caption{\textbf{Ablation study on unreasonable instruction rejection.} With counterfactual behavior generation, the system effectively filters out unreasonable instructions.}
\vspace{-0.5em}
\label{tab:ablation_unreasonable}
\resizebox{0.6\linewidth}{!}{%
\begin{tabular}{cc}
\toprule
\begin{tabular}[c]{@{}c@{}}Counterfactual Behavior\\Generation\end{tabular} & \begin{tabular}[c]{@{}c@{}}Rejection Rate of \\ Unreasonable Instructions (\%) \end{tabular} \\
\midrule
\xmark & 0.00 \\
\cmark & \textbf{93.33} \\
\bottomrule
\end{tabular}}
\vspace{-1.0em}
\end{table}
\vspace{-1.5em}

\subsection{Ours (Coarse) vs. ChatSim}
\label{sec:ours_coarse}
In ChatSim \cite{wei2024editable}, the final video is generated by simply compositing the background with inserted vehicles, without leveraging a video diffusion model to enhance visual quality. 
To verify that our superior performance stems primarily from our core system design (scene graph and agents) rather than the video diffusion refinement, we remove the video diffusion tool and video refinement agent, and compare the coarse videos (produced by the coarse rendering tool) against ChatSim. 
As shown in Table \ref{tab:ours_coarse_vs_chatsim}, even without video diffusion refinement, our coarse-rendered results still outperform ChatSim.
This demonstrates that our agentic architecture itself is a major contribution, independently capable of producing structurally superior results.

\vspace{-0.5em}
\begin{table}[htbp]
  \centering
  \caption{\textbf{Ours (Coarse) vs. ChatSim.} Even without video diffusion refinement, our coarse-rendered results still outperform ChatSim across all metrics. Abbreviations: App. = Appearance, Beh. = Behavior, Str. = Structure Distance, Col. = Collision, Off. = Off-road.}
  \vspace{-0.5em}
  \label{tab:ours_coarse_vs_chatsim}
  \setlength{\tabcolsep}{1.5pt}
  \footnotesize
  \resizebox{0.88\linewidth}{!}{%
  \begin{tabular}{l
                  cc      
                  cc      
                  c       
                  cc}     
    \toprule
    & \multicolumn{2}{c}{Photorealism}
    & \multicolumn{2}{c}{Instr. Align.}
    & \multicolumn{1}{c}{Struct. Pres.}
    & \multicolumn{2}{c}{Traffic Realism} \\
    \cmidrule(r){2-3}\cmidrule(r){4-5}\cmidrule(r){6-6}\cmidrule(r){7-8}
    Method
      & FID $\downarrow$ & FVD $\downarrow$
      & App. (\%) $\uparrow$ & Beh. (\%) $\uparrow$
      & Str. $\downarrow$
      & Col. (\%) $\downarrow$ & Off. (\%) $\downarrow$ \\
    \midrule
    ChatSim \cite{wei2024editable}
      & 47.70  & 605.69
      & 42.33  & 26.64
      & 46.52
      & 27.62 & 24.71 \\
    \textbf{Ours (Coarse)}
      & \textbf{38.19} & \textbf{589.41}
      & \textbf{85.76} & \textbf{73.18}
      & \textbf{37.58}
      & \textbf{3.05}  & \textbf{3.67} \\
    \bottomrule
  \end{tabular}}
  \vspace{-1.0em}
\end{table}

\begin{figure*}[t]
    \centering
    \includegraphics[width=\textwidth]{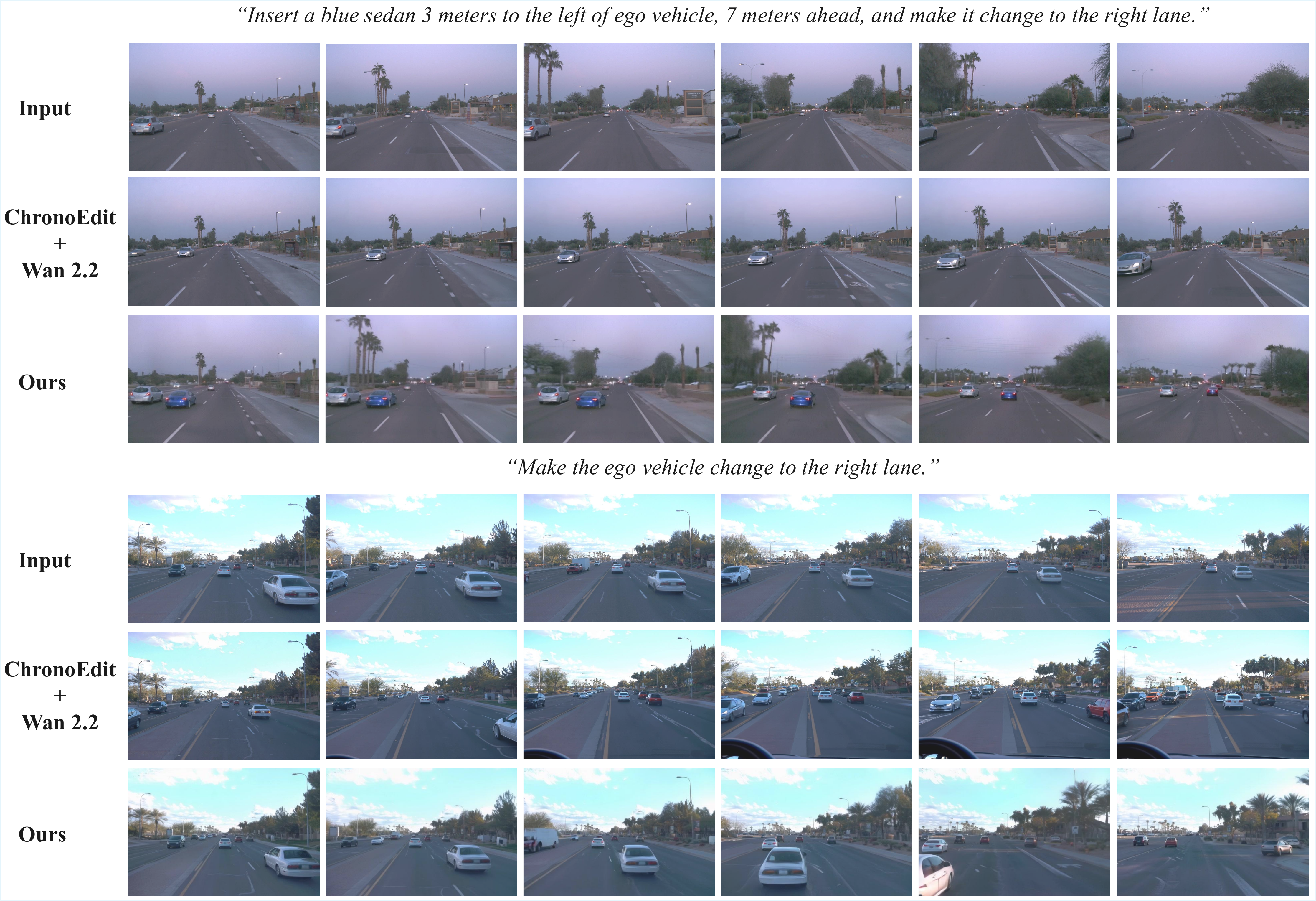}
    \caption{\textbf{Qualitative comparison with image editing + image-to-video methods.} Our method faithfully follows user instructions while preserving the original video background. In contrast, the image editing + image-to-video baseline fails to follow instructions — in the first example, it does not correctly insert the blue sedan, and in the second example, it fails to generate a trajectory matching the target behavior. Furthermore, it also alters the original background, e.g., in the second example, it inserts some vehicles that do not exist in the original scene.}
    \label{fig:supp_ours_wan}
    \vspace{-1em}
\end{figure*}

\subsection{Ours vs. Image editing + Image-to-Video methods}
\label{sec:chronoedit_wan}
We further construct a baseline by naively combining an image editing method with an image-to-video method. Specifically, we apply ChronoEdit~\cite{wu2025chronoedit} to edit the first frame, and then use Wan~2.2~\cite{wan2025wan} to generate a video from the edited frame, with both stages conditioned on the instruction.

As shown in Table \ref{tab:ours_vs_chrono}, our method outperforms the ChronoEdit + Wan~2.2 baseline across all metrics, with qualitative comparisons provided in Figure \ref{fig:supp_ours_wan}. 
Similar to Cosmos \cite{ali2025world}, this baseline suffers from two main issues: 1) the background of the original video is easily altered, as only the first frame is used as input; and 2) it struggles to follow user instructions correctly, failing to accurately remove, replace, or insert the specified vehicles or generate trajectories that match the target behavior.

\vspace{-1.0em}
\begin{table}[htbp]
  \centering
    \caption{\textbf{Ours vs. Image editing + Image-to-Video methods.} Our method outperforms the ChronoEdit + Wan~2.2 baseline across all metrics. Abbreviations: App. = Appearance, Beh. = Behavior, Str. = Structure Distance, Col. = Collision, Off. = Off-road.}
    \vspace{-0.5em}
  \label{tab:ours_vs_chrono}
  \setlength{\tabcolsep}{1.5pt}
  \footnotesize
  \resizebox{0.92\linewidth}{!}{%
  \begin{tabular}{l
                  cc      
                  cc      
                  c       
                  cc}     
    \toprule
    & \multicolumn{2}{c}{Photorealism}
    & \multicolumn{2}{c}{Instr. Align.}
    & \multicolumn{1}{c}{Struct. Pres.}
    & \multicolumn{2}{c}{Traffic Realism} \\
    \cmidrule(r){2-3}\cmidrule(r){4-5}\cmidrule(r){6-6}\cmidrule(r){7-8}
    Method
      & FID $\downarrow$ & FVD $\downarrow$
      & App. (\%) $\uparrow$ & Beh. (\%) $\uparrow$
      & Str. $\downarrow$
      & Col. (\%) $\downarrow$ & Off. (\%) $\downarrow$ \\
    \midrule
    ChronoEdit + Wan 2.2
      & 33.71  & 762.04
      & 51.83  & 39.35
      & 77.37
      & 3.30 & 4.26 \\
    \textbf{Ours}
      & \textbf{32.85} & \textbf{467.20}
      & \textbf{82.19} & \textbf{71.67}
      & \textbf{34.62}
      & \textbf{0.58}  & \textbf{1.73} \\
    \bottomrule
  \end{tabular}}
\end{table}

\subsection{Downstream Task}
We conduct a 3D object detection study using BEVFormer \cite{li2024bevformer} on the Waymo Open Dataset \cite{sun2020scalability}. 
We first prepare 8000 real training frames and then edit them to generate an additional 8000 frames. 
Specifically, we replace existing vehicles in the scene with newly inserted ones. 
As shown in Table \ref{tab:supp_downstream}, augmenting the training set with these edited images consistently improves BEVFormer’s performance across all IoU thresholds, indicating that our edited data is beneficial for downstream tasks.

\begin{table}[!htbp]
\centering
\caption{\textbf{Downstream Task: Object Detection.} With additional edited images as training data, BEVFormer's performance improves consistently across all IoU thresholds.}
\vspace{-0.5em}
\resizebox{0.5\linewidth}{!}{
\begin{tabular}{lccc}
\toprule
Training Data & AP@0.3 $\uparrow$ & AP@0.5 $\uparrow$ & AP@0.7 $\uparrow$ \\
\midrule
Real        & 0.1211 & 0.0533 & 0.0103 \\
\textbf{Real + Edited}  & \textbf{0.1343} & \textbf{0.0635} & \textbf{0.0125} \\
\bottomrule
\end{tabular}}
\label{tab:supp_downstream}
\vspace{-1em}
\end{table}

\subsection{Cross-Model Judge Agreement Validation}
In Table \ref{tab:base_exp} of the main paper, we use GPT-4o \cite{hurst2024gpt} to evaluate the appearance alignment and traffic realism metrics. To further validate the consistency across different judge models, we additionally introduce Claude-Sonnet-4.6 \cite{anthropic2026claudesonnet46} and Gemini-2.5-Pro \cite{comanici2025gemini} as evaluators. As shown in Table \ref{tab:judge_models}, our method consistently outperforms the baselines across all judge models.

\begin{table}[htbp]
\vspace{-1em}
  \centering
  \caption{\textbf{Cross-model judge agreement validation on appearance alignment, collision rate, and off-road rate.} We compare GPT-4o, Gemini-2.5-Pro, and Claude-Sonnet-4.6 as judge models. Our method consistently outperforms the baselines across all judge models.}
  \setlength{\tabcolsep}{3pt}
  \renewcommand{\arraystretch}{1.05}
  \scriptsize
  \resizebox{0.92\columnwidth}{!}{%
  \begin{tabular}{l ccc ccc ccc}
    \toprule
    &
    \multicolumn{3}{c}{App. Align. (\%) $\uparrow$}
    & \multicolumn{3}{c}{Collision (\%) $\downarrow$}
    & \multicolumn{3}{c}{Off-road (\%) $\downarrow$} \\
    \cmidrule(r){2-4}\cmidrule(r){5-7}\cmidrule(r){8-10}
    Method
    & GPT-4o & Gemini & Claude
    & GPT-4o & Gemini & Claude
    & GPT-4o & Gemini & Claude \\
    \midrule
    Cosmos \cite{ali2025world}
      & 46.25 & 44.60 & 43.17
      & 3.16 & 5.22 & 1.97
      & 3.95 & 5.49 & 1.26 \\
    ChatSim \cite{wei2024editable}
      & 42.33 & 39.87 & 45.63
      & 27.62 & 27.03 & 23.46
      & 24.71 & 24.26 & 21.38 \\
    \textbf{Ours}
      & \textbf{82.19} & \textbf{78.43} & \textbf{80.95}
      & \textbf{0.58} & \textbf{1.98} & \textbf{0.16}
      & \textbf{1.73} & \textbf{3.36} & \textbf{0.45} \\
    \bottomrule
  \end{tabular}}
  \label{tab:judge_models}
  \vspace{-2em}
\end{table}

\section{Detailed Comparison with Related Work}

\begin{table*}[t]
\centering
\small  
\caption{\textbf{Detailed comparison with related work.} Our method supports the most editing operations and achieves the best editing results.}
\label{tab:comparison}
\resizebox{\textwidth}{!}{  
\begin{tabular}{lcccccccccccc}
\toprule
& \multicolumn{7}{c}{Editing Capacities} & \multicolumn{4}{c}{Editing Performance} \\
\cmidrule(lr){2-8} \cmidrule(lr){9-12}
& \begin{tabular}[c]{@{}c@{}}Open- \\Vocabulary\\Object Query\end{tabular} & \begin{tabular}[c]{@{}c@{}}Object\\Removal\end{tabular} & \begin{tabular}[c]{@{}c@{}}Object\\Insertion\end{tabular} & \begin{tabular}[c]{@{}c@{}}Object\\Replacement\end{tabular} & \begin{tabular}[c]{@{}c@{}}Trajectory\\Editing\end{tabular} & \begin{tabular}[c]{@{}c@{}}Ego Vehicle\\View Editing\end{tabular} & \begin{tabular}[c]{@{}c@{}}Multi-object\\Simulation\end{tabular} & \begin{tabular}[c]{@{}c@{}}Photo-\\realism\end{tabular} & \begin{tabular}[c]{@{}c@{}}Instruction\\Alignment\end{tabular} & \begin{tabular}[c]{@{}c@{}}Structure\\Preservation\end{tabular} & \begin{tabular}[c]{@{}c@{}}Traffic\\Realism\end{tabular} \\
\midrule
\multicolumn{12}{l}{\textit{Diffusion-based Methods}} \\
DriveEditor \cite{liang2025driveeditor} & \xmark & \cmark & \cmark & \cmark & \xmark & \xmark & \xmark & \cmark & \xmark & \cmark & \cmark \\
SceneCrafter \cite{zhu2025scenecrafter} & \xmark & \cmark & \cmark & \cmark & \xmark & \xmark & \xmark & \cmark & \xmark & \cmark & \xmark \\
Cosmos \cite{ali2025world} & \cmark & \cmark & \cmark & \cmark & \cmark & \cmark & \xmark & \cmark & \xmark & \xmark & \cmark \\
\midrule
\multicolumn{12}{l}{\textit{Gaussian Splatting-based Methods}} \\
OminiRe \cite{chen2024omnire} & \xmark & \cmark & \cmark & \cmark & \xmark & \xmark & \xmark & \xmark & \xmark & \cmark & \xmark \\
DrivingGaussian++ \cite{xiong2025drivinggaussian++} & \xmark & \cmark & \cmark & \cmark & \cmark & \xmark & \xmark & \xmark & \xmark & \cmark & \xmark \\
\midrule
\multicolumn{12}{l}{\textit{Agent-based Methods}} \\
AutoVFX \cite{hsu2025autovfx} & \cmark & \cmark & \cmark & \cmark & \xmark & \xmark & \xmark & \xmark & \xmark & \cmark & \xmark \\
ChatDyn \cite{wei2024chatdyn} & \cmark & \cmark & \cmark & \cmark & \cmark & \xmark & \cmark & \xmark & \xmark & \cmark & \cmark \\
ChatSim \cite{wei2024editable} & \cmark & \cmark & \cmark & \cmark & \cmark & \cmark & \xmark & \xmark & \xmark & \cmark & \xmark \\
\midrule
\textbf{Ours} & \cmark & \cmark & \cmark & \cmark & \cmark & \cmark & \cmark & \cmark & \cmark & \cmark & \cmark \\
\bottomrule
\end{tabular}
}  
\vspace{-2em}
\end{table*}

We provide a detailed comparison with previous driving scene editing methods in Table~\ref{tab:comparison}.
Prior work can be roughly grouped into three categories.

The first category consists of diffusion-based methods \cite{liang2025driveeditor,zhu2025scenecrafter,ali2025world}. 
Among them, DriveEditor \cite{liang2025driveeditor} and SceneCrafter \cite{zhu2025scenecrafter} do not support open-vocabulary object query and require the user to specify the edited region via 2D/3D bounding boxes. 
Moreover, these methods struggle to generate realistic target trajectories and model multi-object interactions.
The second category includes Gaussian Splatting-based methods \cite{chen2024omnire,xiong2025drivinggaussian++}. 
Similarly, they lack open-vocabulary object query capability and require manual selection of target objects. 
Moreover, their editing results exhibit poor photorealism and traffic realism.
The third category consists of LLM \cite{achiam2023gpt,hurst2024gpt,touvron2023llama,bai2023qwen,zheng2025parallel,zheng2026parallel} agent-based pipelines \cite{hsu2025autovfx,wei2024chatdyn,wei2024editable}.
While these methods enable purely natural language-based editing, their generated results often exhibit inconsistent lighting between newly inserted objects and the original background, resulting in visually unnatural appearances. 
Additionally, the generated trajectories are not realistic.
Furthermore, SimSplat~\cite{park2025simsplat} is a concurrent work that also performs editing based on scene graph representation and uses an agent-based framework. 
However, unlike our approach, it does not incorporate iterative behavior and video refinement modules to enhance photorealism, instruction alignment and traffic realism.

In Table \ref{tab:base_exp} of the main paper, we therefore compare our method against the state-of-the-art open-source models Cosmos \cite{ali2025world} and ChatSim \cite{wei2024editable}, both of which support purely natural language-based editing.

\section{Implementation Details}
\subsection{Counterfactual Behavior Generation and Behavior Validation}
\label{sec:behavior_description_and_validation}

Building upon \cite{chang2025langtraj}, we extract semantic behavior descriptions from original object trajectory and introduce a novel automated engine for reasoning about the physical and semantic consistency of counterfactual behaviors. These technologies are leveraged by the Object Grounding, Behavior Editing, and Behavior Reviewer Agents.

\subsubsection{Behavior Description Generation}

We define an object's state sequence as $S = \{s_1, \dots, s_T\}$ and the local vector map as $\mathcal{M}$. The map is parsed to construct a connectivity graph $\mathcal{L}$, where intersections are inferred via density-based spatial clustering (DBSCAN) \cite{ester1996density} of lane centerline conflict points. For each object, we extract a set of ground truth behavior tokens $\mathcal{A}_{gt}$ (behavior descriptions of the original trajectory) using the following geometric and kinematic primitives.

\paragraph{Kinematic State Classification.}
We classify longitudinal motion by analyzing the object's displacement derivatives. 
To account for sensor noise, we employ adaptive thresholds. 
An object is classified as \textit{static} if total displacement $< 0.5$m. 
For moving objects, speed patterns are categorized as \textit{speeding up}, \textit{slowing down}, or \textit{varying speed} based on the monotonicity of velocity changes ($\Delta v$) over a smoothing window, subject to a relaxation parameter $\epsilon$ to allow for minor fluctuations.

\paragraph{Map-Adaptive Topology.}
Lateral behaviors are determined by projecting the object's position onto $\mathcal{L}$. 
We assign lane ownership (e.g., \textit{in leftmost lane}) by computing the nearest lane centerline with a heading alignment tolerance of $\pm 10^\circ$. Lane change maneuvers are identified when an object transitions between adjacent lane IDs over a duration threshold $t_{lc} \geq 3$ frames, provided the lanes are not topological successors.

\paragraph{Intersection Interaction.}
We model intersections as buffered regions around the centroids of clustered lane conflicts. Complex maneuvers are inferred via geometric triggers:
\begin{itemize}
    \item \textbf{Approaching:} The object is within a look-ahead distance of an intersection centroid and maintains a velocity $v < v_{safe}$, where $v_{safe}$ is the maximum cornering speed derived from a friction circle model.
    \item \textbf{Crossing:} 
    The object's trajectory physically intersects the polygon buffer of an intersection.
    \item \textbf{Turning:} We integrate the cumulative heading change $\Delta \theta$. The object is assigned \textit{turning left} if $\sum \Delta \theta > \pi/6$, \textit{turning right} if $\sum \Delta \theta < -\pi/6$, and \textit{going straight} otherwise.
\end{itemize}

\subsubsection{Counterfactual Behavior Generation}

To capture the multimodality of driving scenes, we develop a novel method to generate the set of physically and semantically consistent counterfactual actions $\mathcal{A}_{cf}$ — behaviors the object \textit{could} have executed but did not. 
The synthesis pipeline operates in three stages:

\paragraph{1). Token-Level Expansion.}
We define a mapping function $\Phi: t \rightarrow \{c_1, \dots, c_n, \emptyset\}$ that maps an observed ground truth behavior token $t$ to a set of plausible alternatives. The complete mapping logic, derived from kinematic feasibility, is detailed in Table~\ref{tab:full_counterfactuals}. Note that we explicitly include a null token ($\emptyset$) to allow the model to generate simplified descriptions by ``forgetting'' specific details (e.g., removing speed information).

\paragraph{2). Combinatorial Generation.}
We generate the candidate space $\mathbb{S}$ via the Cartesian product of the token choices:
\begin{equation}
    \mathbb{S} = \prod_{t \in \mathcal{A}_{gt}} (\{t\} \cup \Phi(t))
\end{equation}
This expansion produces a dense set of potential behaviors, many of which may be physically impossible (e.g., \textit{static} combined with \textit{changing lanes}).

\paragraph{3). Semantic Compatibility Pruning.}
To ensure physical consistency, we enforce a compatibility matrix $\mathcal{C}$. A candidate description $S_{cand} \in \mathbb{S}$ is valid if and only if:
\begin{equation}
    \forall a_i, a_j \in S_{cand}, \quad a_i \notin \text{Incompatible}(a_j)
\end{equation}
The incompatibility constraints are detailed in Table~\ref{tab:full_incompatibilities}. We specifically enforce that \textit{static} and \textit{parked} states are mutually exclusive with all behavior tokens. 
Additionally, we apply context-aware filtering to prune lane changing and turning hallucinations that violate the map topology (e.g., removing \textit{change to the left lane} if the object is already in the \textit{leftmost lane}, removing \textit{turn left} if there is \textit{no intersection}). Finally, strict subset behaviors are pruned to prioritize maximal specificity.

\subsubsection{Behavior Validation} 
The Behavior Reviewer Agent uses the same logic as in behavior description generation to determine if generated trajectories align with target behaviors. It also checks if generated trajectories contain traffic violations, i.e., off-road behavior and collisions.
Off-road behavior is identified when the majority of trajectory points lie outside road boundaries. 
For collision detection, each vehicle is first represented as an oriented bounding box. Then at each time step, the Separating Axis Theorem \cite{gottschalk1996obbtree} is used to detect overlaps between vehicles for collision checking.

\subsubsection{Behavior Alignment Metric}
In Table \ref{tab:base_exp} of the main paper, we calculate the behavior alignment metric using the same logic as in behavior description generation.
Although our method generates explicit trajectories during the editing process, we do not use them directly for evaluation. 
Instead, to ensure fair comparison with  baselines, we apply the same evaluation protocol to all methods: first use the tracking model \cite{ren2024grounded} to track vehicles in edited videos and then transform the tracked trajectories to world coordinates for evaluation.

\subsection{Test Dataset}
In Table \ref{tab:test_scenes}, we list the IDs of 30 test scenes selected from the Waymo Open Dataset \cite{sun2020scalability}, which cover different times of day, weather conditions, and road types.
For all test scenes, we use the first 80 frames from the front camera as the input video.

\begin{table}[h]
\centering
\footnotesize
\caption{Test scene IDs selected from the Waymo Open Dataset \cite{sun2020scalability}.}
\label{tab:test_scenes}
\begin{tabular}{l}
\toprule
\multicolumn{1}{c}{Scene ID} \\
\midrule
segment-1005081002024129653\_5313\_150\_5333\_150 \\
segment-10923963890428322967\_1445\_000\_1465\_000 \\
segment-10927752430968246422\_4940\_000\_4960\_000 \\
segment-11839652018869852123\_2565\_000\_2585\_000 \\
segment-14940138913070850675\_5755\_330\_5775\_330 \\
segment-15803855782190483017\_1060\_000\_1080\_000 \\
segment-16552287303455735122\_7587\_380\_7607\_380 \\
segment-16651261238721788858\_2365\_000\_2385\_000 \\
segment-2273990870973289942\_4009\_680\_4029\_680 \\
segment-3338044015505973232\_1804\_490\_1824\_490 \\
segment-3665329186611360820\_2329\_010\_2349\_010 \\
segment-4537254579383578009\_3820\_000\_3840\_000 \\
segment-5076950993715916459\_3265\_000\_3285\_000 \\
segment-6150191934425217908\_2747\_800\_2767\_800 \\
segment-6207195415812436731\_805\_000\_825\_000 \\
segment-6935841224766931310\_2770\_310\_2790\_310 \\
segment-10335539493577748957\_1372\_870\_1392\_870 \\
segment-11660186733224028707\_420\_000\_440\_000 \\
segment-12496433400137459534\_120\_000\_140\_000 \\
segment-12820461091157089924\_5202\_916\_5222\_916 \\
segment-13299463771883949918\_4240\_000\_4260\_000 \\
segment-15021599536622641101\_556\_150\_576\_150 \\
segment-15056989815719433321\_1186\_773\_1206\_773 \\
segment-16229547658178627464\_380\_000\_400\_000 \\
segment-16767575238225610271\_5185\_000\_5205\_000 \\
segment-16979882728032305374\_2719\_000\_2739\_000 \\
segment-17152649515605309595\_3440\_000\_3460\_000 \\
segment-25067997087482581165\_6455\_000\_6475\_000 \\
segment-45753894051788059994\_4900\_000\_4920\_000 \\
segment-53722817286274376181\_2005\_000\_2025\_000 \\
\bottomrule
\end{tabular}
\vspace{-2em}
\end{table}

\subsection{Agent Details}
In this section, we present the detailed reasoning process of each agent, including the specific instructions and prompts.

Figure \ref{fig:orchestrator} illustrates the orchestrator's workflow. 
We provide the orchestrator with predefined functions and templates of the complete editing workflow. Additionally, we include examples that map user instructions to their corresponding Python code.
During inference, this enables the orchestrator to automatically generate executable scripts based on user instructions.

Figure \ref{fig:object_grounding_agent} demonstrates the process employed by the object grounding agent. 
The agent first decomposes textual descriptions into triplets of (reference object, direction, target object). 
It then identifies the best-matching reference object using appearance, behavior, and position information. 
After filtering candidates by directional constraints, it applies the same matching procedure to locate the target object.

Figure \ref{fig:insertion_agent} presents the insertion agent's pipeline. 
The agent first estimates the real-world size of the object from its textual description, then computes scaling factors by comparing it to the generated mesh dimensions.
Next, it determines the mesh's local coordinate system by analyzing the object's orientation in rendered images. Based on this, it derives the transformation matrix from local coordinate system to the scene's world coordinate system.

Figure \ref{fig:behavior_editing_agent} shows the counterfactual behavior selection process within the behavior editing agent. 
This component selects the behavior combination from available counterfactuals that most closely aligns with the target behavior, while also preserving as much of the original behavior as possible.

Figure \ref{fig:behavior_reviewer_agent} illustrates the workflow of the behavior reviewer agent. 
Based on validation results from the generated trajectories, the agent adjusts the guidance mode and its corresponding configuration accordingly.

Figure~\ref{fig:video_reviewer_agent} illustrates the pipeline of the video reviewer agent. It first localizes the inserted vehicles using their masks. It then compares the corresponding regions in the coarse and refined video frames to assess two aspects: 1) whether the inserted vehicles appear realistic in the refined frame, e.g., whether their lighting is consistent with the surrounding environment; and 2) whether the appearance of the inserted vehicles is preserved. If the vehicles appear unrealistic, the agent increases the denoising strength $\sigma$; if appearance is not preserved, it increases the L2 guidance loss weight $\lambda$.

Formally, the denoising strength $s \in (0, 1]$ controls the global editing magnitude by determining the starting timestep of the diffusion process. Given the total number of denoising steps $N$, the number of active denoising steps $K$ and the starting index $t_{\text{start}}$ are:
\begin{equation}
    K = \min(\text{int}(N \cdot s), N), \quad t_{\text{start}} = N - K,
    \label{eq:strength}
\end{equation}
where $t_{\text{start}}$ is the index into the scheduler's \cite{song2020denoising} timestep sequence, and the corresponding actual starting timestep is $t_0 = \text{scheduler.timesteps}[t_{\text{start}}]$. A larger $s$ results in more denoising steps, producing more photorealistic outputs at the cost of reduced appearance consistency. The initial latent is obtained by adding noise to the condition video latent \cite{kingma2013auto} $\mathbf{x}_{\text{ref}}$ at the starting timestep $t_0$:
\begin{equation}
    \mathbf{x}_{t_0} = \sqrt{\bar{\alpha}_{t_0}}\, \mathbf{x}_{\text{ref}} + \sqrt{1 - \bar{\alpha}_{t_0}}\, \mathbf{z}, \quad \mathbf{z} \sim \mathcal{N}(\mathbf{0}, \mathbf{I}),
    \label{eq:init_latent}
\end{equation}
where $\bar{\alpha}_{t_0}$ is the cumulative noise schedule coefficient at $t_0$. At each denoising step $t$, we first apply classifier-free guidance (CFG) to obtain the guided noise prediction:
\begin{equation}
    \boldsymbol{\epsilon}_{\text{cfg}} = \boldsymbol{\epsilon}_u + w(\boldsymbol{\epsilon}_c - \boldsymbol{\epsilon}_u),
    \label{eq:cfg}
\end{equation}
where $\boldsymbol{\epsilon}_u$ and $\boldsymbol{\epsilon}_c$ are the unconditional and conditional noise predictions respectively, and $w$ is the CFG guidance scale. We then estimate the predicted clean latent $\hat{\mathbf{x}}_0$ from the current noisy latent $\mathbf{x}_t$:
\begin{equation}
    \hat{\mathbf{x}}_0 = \frac{\mathbf{x}_t - \sqrt{1 - \bar{\alpha}_t}\, \boldsymbol{\epsilon}_{\text{cfg}}}{\sqrt{\bar{\alpha}_t}}.
    \label{eq:pred_x0}
\end{equation}
To preserve the appearance of inserted vehicles, we compute the masked residual between the predicted clean latent and the condition video latent over the vehicle regions:
\begin{equation}
    \mathbf{e}_t = \mathcal{M} \odot \left( \hat{\mathbf{x}}_0 - \mathbf{x}_{\text{ref}} \right),
    \label{eq:residual}
\end{equation}
where $\mathcal{M} \in \{0,1\}^{H \times W}$ is the binary mask of the inserted-vehicle regions resized to the latent resolution, and $\odot$ denotes element-wise multiplication. The L2 guidance loss is then defined as:
\begin{equation}
    \mathcal{L}_{\text{L2}} = \left\| \mathbf{e}_t \right\|_2^2,
    \label{eq:l2_loss}
\end{equation}
which is incorporated into the noise prediction by injecting the scaled residual into the noise space:
\begin{equation}
    \tilde{\boldsymbol{\epsilon}}_t = \boldsymbol{\epsilon}_{\text{cfg}} + \lambda_t \cdot \mathbf{e}_t, \quad \lambda_t = \lambda \cdot \frac{\sqrt{\bar{\alpha}_t}}{\sqrt{1 - \bar{\alpha}_t} + 10^{-8}},
    \label{eq:guidance}
\end{equation}
where $\lambda \geq 0$ is the L2 guidance weight controlling the strength of appearance preservation. The latent is then updated to the next timestep via the DDIM scheduler~\cite{song2020denoising}:
\begin{equation}
    \mathbf{x}_{t-1} = \text{SchedulerStep}(\tilde{\boldsymbol{\epsilon}}_t, t, \mathbf{x}_t).
    \label{eq:scheduler}
\end{equation}
In summary, the denoising strength $s$ determines the global editing range by controlling how many denoising steps to perform, while the L2 guidance weight $\lambda$ enforces local appearance preservation at each step by pulling the predicted latent toward the condition video latent. Based on the review, the agent dynamically adjusts $s$ and $\lambda$ at each iteration to jointly optimize photorealism and appearance preservation.

\section{Extra Qualitative Results}
\begin{figure*}[htbp]
    \vspace{-0.2cm}
    \centering
    \includegraphics[width=0.96\textwidth]{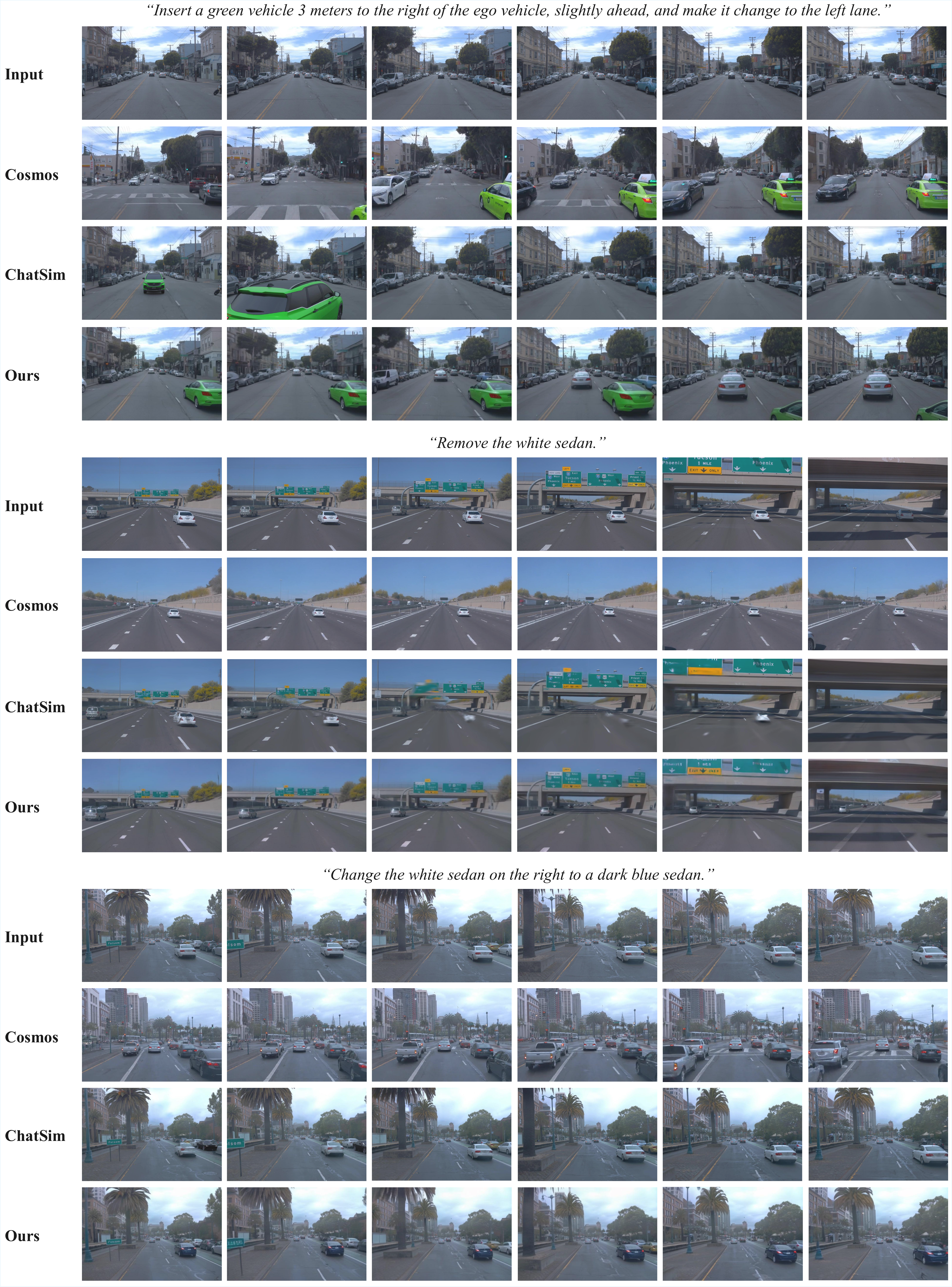}
    \caption{\textbf{Extra qualitative comparison with baselines.} Our method significantly outperforms previous approaches across all four aspects: photorealism, instruction alignment, structure preservation, and traffic realism. Note in the first instruction, when the newly inserted green sedan cuts in, both the ego vehicle and the green sedan recognize they are too close and decide to stop, which demonstrates that our method can effectively simulate safety-critical long-tail scenarios.}
    \label{fig:supp_base}
\end{figure*}

\begin{figure*}[htbp]
    \centering
    \includegraphics[width=\textwidth]{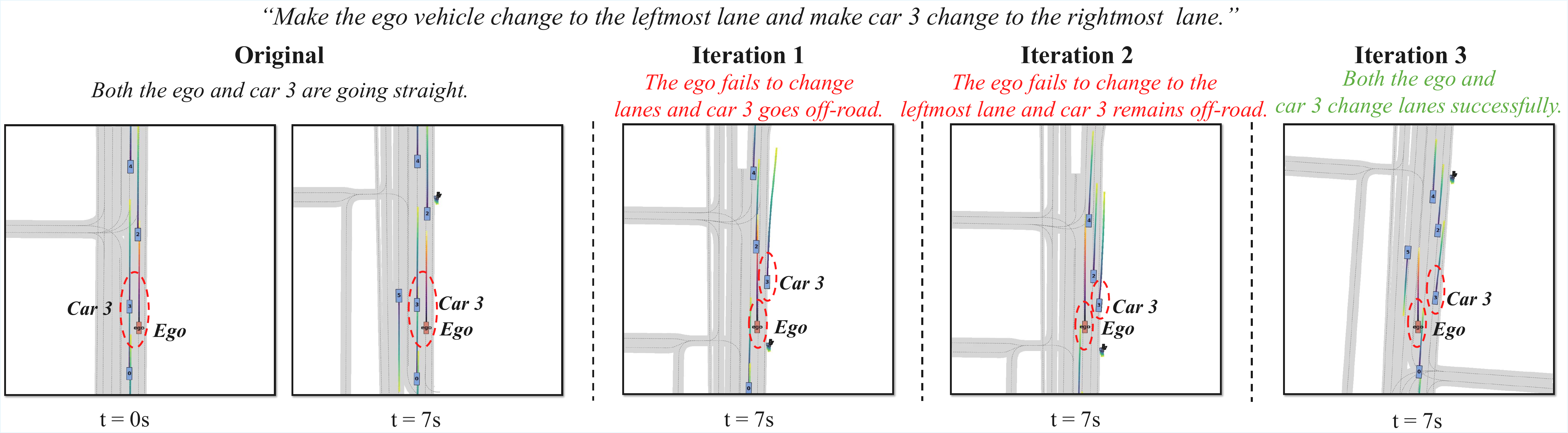}
    \vspace{-2em}
    \caption{\textbf{Extra qualitative results for behavior feedback loop.} In iteration 1, the ego vehicle remains in its lane, while car 3 changes lanes but goes off-road. 
    The reviewer agent then increases the classifier-free guidance weight for the ego vehicle and applies on-road guidance to car 3. 
    In iteration 2, the ego vehicle changes lanes but fails to reach the leftmost lane in its direction, while part of car 3's trajectory remains off-road. 
    So the reviewer agent further increases both guidance weights. 
    In iteration 3, both vehicles successfully generate valid trajectories. 
    (Note: leftmost and rightmost lanes refer to lanes within the same traffic direction.)}
    \vspace{-1em}
    \label{fig:supp_behavior_feedback_loop}
\end{figure*}

\begin{figure*}[htbp]
    \centering
    \includegraphics[width=0.98\textwidth]{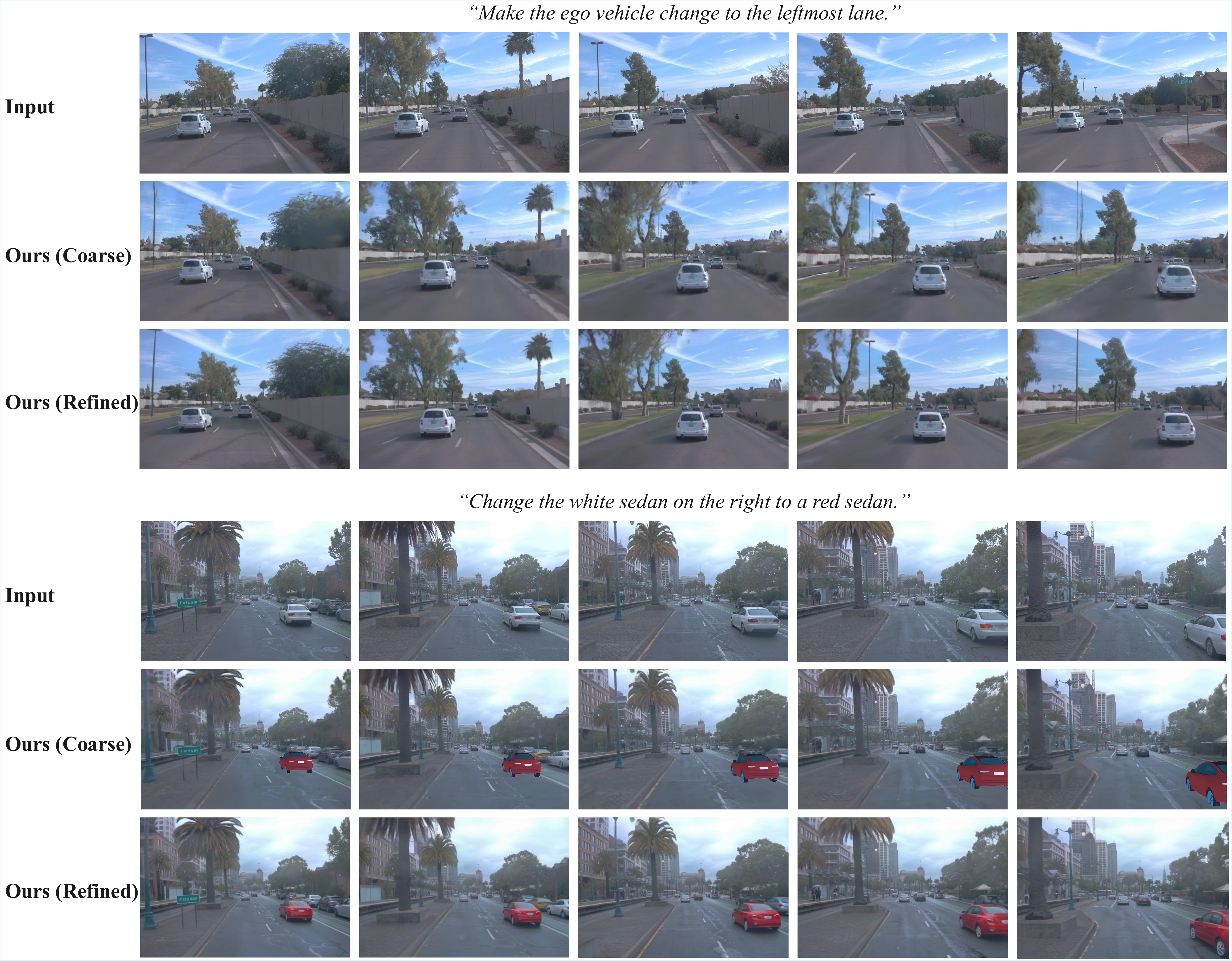}
    \caption{\textbf{Qualitative results before and after the iterative video diffusion refinement.}
    ``Ours (Coarse)'' refers to the coarse video produced by the coarse rendering tool, while ``Ours (Refined)'' refers to the video refined by the video diffusion tool and video reviewer agent.
    Typically, visual quality suffers in two scenarios: 1) when viewpoints change substantially, rendering quality drops significantly; 2) when new objects are inserted, meshes appear inconsistent with the original scene. The iterative video diffusion refinement effectively addresses both issues.}
    \label{fig:supp_vdm}
\end{figure*}

\begin{figure*}[htbp]
    \centering
    \includegraphics[width=0.65\textwidth]{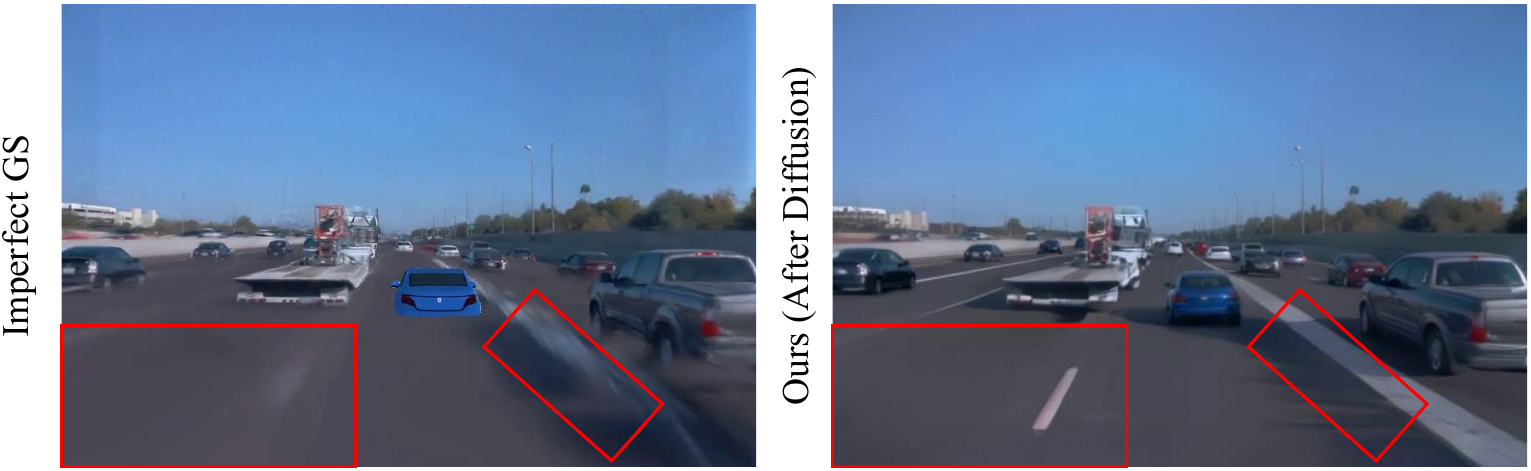}
    \caption{\textbf{Robustness to Gaussian reconstruction artifacts.} Imperfect Gaussian splats introduce visible artifacts on the road surface and lane markings, as highlighted in red. Our video diffusion tool effectively removes these artifacts while preserving realistic scene appearance and the inserted vehicle.}
    \label{fig:imperfect_gs}
\end{figure*}

In this section, we provide additional qualitative results. 
Specifically, Figure~\ref{fig:supp_base} shows editing results of different methods across various instruction types. 
As observed, Cosmos \cite{ali2025world} modifies the original background, while ChatSim \cite{wei2024editable} suffers from poor photorealism. Moreover, neither method follows instructions well (e.g., the initial position and behavior of newly added vehicles). 
Additionally, we observe an interesting phenomenon in the insertion example (``Insert a green vehicle 3 meters to the right of the ego vehicle, slightly ahead, and make it change to the left lane."). When the newly inserted green sedan cuts in, both the ego vehicle and the green sedan recognize they are too close and decide to stop. 
This demonstrates that our method can effectively simulate safety-critical long-tail scenarios.
Figure~\ref{fig:supp_behavior_feedback_loop} visualizes the effect of the behavior feedback loop. 
By adjusting the guidance configuration based on the behavior validation results, the agent generates trajectories that match the target behavior while avoiding collisions and off-road violations.
Figure~\ref{fig:supp_vdm} presents a visual comparison before and after iterative video diffusion refinement. 
As shown, the refined videos not only significantly improve visual quality (addressing rendering quality degradation caused by ego-viewpoint changes and ensuring lighting and style consistency between inserted objects and the environment), but also preserve the appearance of inserted vehicles.
Figure~\ref{fig:imperfect_gs} shows that our method is highly robust to reconstruction artifacts. 
To reduce failures caused by imperfect Gaussian reconstruction, we train the video diffusion tool with a cycle-reconstruction strategy \cite{wu2025difix3d+}. Specifically, we first render frames from the reconstructed Gaussian splats under perturbed trajectories. We then reconstruct a new Gaussian field from these perturbed-view renderings and render it back under the original trajectory to obtain noisy-clean training pairs.

\begin{figure*}[htbp]
    \centering
    \includegraphics[width=\textwidth]{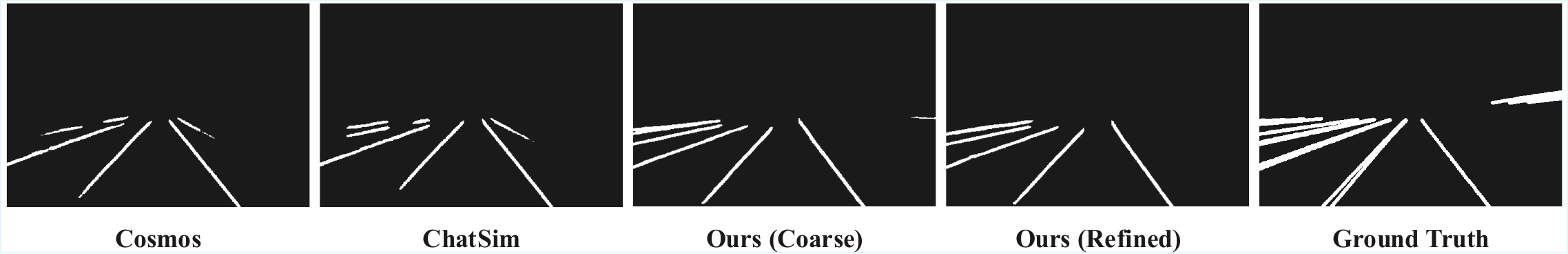}
    \caption{\textbf{Lane preservation comparison across different methods.} For the ground truth, we directly project the 3D lane from the map into pixel space. 
    For all other methods, we detect lanes in the generated videos using the lane detection model TwinLiteNet \cite{che2023twinlitenet}. 
    Here, ``Ours (Coarse)'' refers to the video produced by the coarse rendering tool, while ``Ours (Refined)'' refers to the video after refinement by the video diffusion model. 
    Although the video diffusion model slightly alters the lane structure, our method still significantly outperforms the baselines.}
    \label{fig:lane}
\end{figure*}

Additionally, Table \ref{tab:hallucination_analysis} of the main paper reports the NTL-IoU metric \cite{zhao2025drivedreamer4d}, which measures how well each method preserves the road structure of the input video. Qualitative results are shown in Figure \ref{fig:lane}. 
For the ground truth, we directly project the 3D lane from the map into pixel space. 
For all other methods, we detect lanes in the generated videos using the lane detection model TwinLiteNet \cite{che2023twinlitenet}. 
Although the video diffusion model slightly alters the lane structure, our method still significantly outperforms the baselines.

\section{Failure Case}

\begin{figure*}[htbp]
    \centering
    \includegraphics[width=\textwidth]{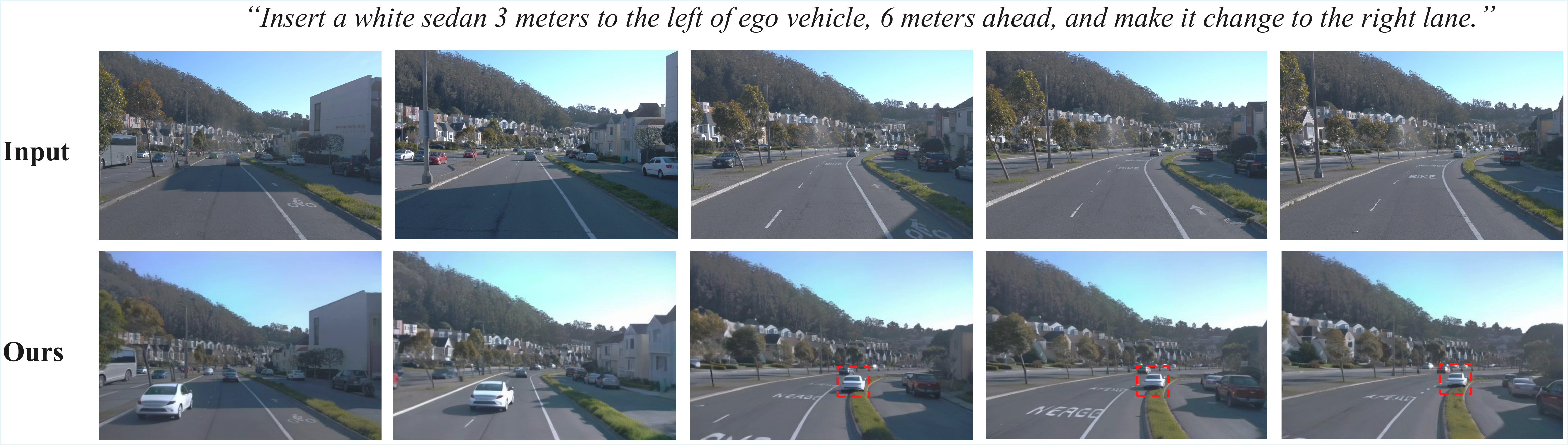}
    \caption{\textbf{Failure case: traffic violation.} The newly inserted vehicle (highlighted in red) incorrectly drives on the median barrier, as the behavior editing module fails to recognize it as a non-drivable area.}
    \label{fig:failure_traj}
\end{figure*}

In this section, we discuss two common failure cases of our method. First, our method may occasionally suffer from object-grounding or localization errors, in which it fails to accurately identify the target object to be edited. We provide a quantitative analysis of this issue in Table \ref{tab:object_grounding}. Second, the generated trajectories may still contain traffic violations, largely due to the difficulty of fine-grained road-structure reasoning and complex multi-vehicle interactions. For example, the system may fail to recognize road separations such as median barriers and instead treat them as drivable areas. As shown in Figure~\ref{fig:failure_traj}, the newly inserted vehicle ends up driving onto the median barrier.

\clearpage  

\begin{table*}[p]  
\centering
\vspace*{\fill} 
\scriptsize
\renewcommand{\arraystretch}{1.2}
\caption{\textbf{Counterfactual Mapping Function $\Phi(t)$.} This table enumerates the set of alternative behavior tokens generated for every observed vehicle behavior. The generation process permutes these tokens to create diverse textual behavior descriptions from a single trajectory. Note: Implicitly, all tokens can also map to $\emptyset$ (DROP).}
\label{tab:full_counterfactuals}
\begin{tabularx}{\linewidth}{@{}l >{\raggedright\arraybackslash}X@{}}
\toprule
\textbf{Observed Behavior ($t$)} & \textbf{Counterfactual Candidates ($\Phi(t)$)} \\ 
\midrule
\multicolumn{2}{l}{\textit{Directional Maneuvers}} \\
\midrule
Going Straight & Turning Left, Turning Right, Slowing Down, Speeding Up \\
Turning Left & Going Straight, Turning Right, Slowing Down \\
Turning Right & Going Straight, Turning Left, Slowing Down \\
Approaching Intersection & Crossing Intersection, Turning Left, Turning Right, Going Straight \\
Crossing Intersection & Approaching Intersection, Turning Left, Turning Right, Going Straight \\
Off Main Roads & Slowing Down, Speeding Up, Turning Left, Turning Right, Going Straight \\
\midrule
\multicolumn{2}{l}{\textit{Longitudinal Dynamics}} \\
\midrule
Speeding Up & Slowing Down, Varying Speed \\
Slowing Down & Speeding Up, Varying Speed \\
Varying Speed & Slowing Down, Speeding Up \\
Moving Slowly & Static, Parked, Off Main Roads, Speeding Up \\
Static & Speeding Up, Moving Slowly \\
Parked & Speeding Up, Moving Slowly \\
\midrule
\multicolumn{2}{l}{\textit{Lane Position}} \\
\midrule
In Leftmost Lane & Changing Lanes (Left $\to$ Mid), Changing Lanes (Left $\to$ Right), Going Straight \\
In Middle Lane & Changing Lanes (Mid $\to$ Left), Changing Lanes (Mid $\to$ Right), Going Straight \\
In Rightmost Lane & Changing Lanes (Right $\to$ Left), Changing Lanes (Right $\to$ Mid), Going Straight \\
\midrule
\multicolumn{2}{l}{\textit{Lane Change Maneuvers}} \\
\midrule
Change: Left $\to$ Mid & In Leftmost Lane, Change (Left $\to$ Right), Change (Mid $\to$ Left), Change (Mid $\to$ Right) \\
Change: Left $\to$ Right & In Leftmost Lane, Change (Left $\to$ Mid), Change (Right $\to$ Left), Change (Right $\to$ Mid) \\
Change: Mid $\to$ Left & In Middle Lane, Change (Mid $\to$ Right), Change (Left $\to$ Mid), Change (Left $\to$ Right) \\
Change: Mid $\to$ Right & In Middle Lane, Change (Mid $\to$ Left), Change (Right $\to$ Left), Change (Right $\to$ Mid) \\
Change: Right $\to$ Left & In Rightmost Lane, Change (Right $\to$ Mid), Change (Left $\to$ Mid), Change (Left $\to$ Right) \\
Change: Right $\to$ Mid & In Rightmost Lane, Change (Right $\to$ Left), Change (Mid $\to$ Left), Change (Mid $\to$ Right) \\
\bottomrule
\end{tabularx}
\vspace*{\fill}  
\end{table*}

\clearpage

\clearpage 

\begin{table*}[p]  
\centering
\vspace*{\fill}  
\scriptsize
\renewcommand{\arraystretch}{1.2}
\caption{\textbf{Compatibility Constraints Matrix ($\mathcal{C}$).} Pruning logic derived from physical and semantic conflicts. 
A generated behavior token is discarded if it contains behavior $A$ along with any behavior from its incompatible set.}
\label{tab:full_incompatibilities}
\begin{tabularx}{\linewidth}{@{}l >{\raggedright\arraybackslash}X@{}}
\toprule
\textbf{Behavior ($A$)} & \textbf{Incompatible Set (Mutually Exclusive with $A$)} \\ 
\midrule
\multicolumn{2}{l}{\textit{Global State}} \\
\midrule
Static & All other behaviors (including Parked, all Moving, all Turning, all Lane behaviors). \\
Parked & All other behaviors (including Static, all Moving, all Turning, all Lane behaviors). \\
Off Main Roads & Static, Crossing Intersection, Approaching Intersection, all Lane behaviors. \\
\midrule
\multicolumn{2}{l}{\textit{Directional Maneuvers}} \\
\midrule
Going Straight & Turning Left, Turning Right, Static, Parked. \\
Turning Left & Going Straight, Turning Right, Crossing Intersection, Approaching Intersection, Static, Parked. \\
Turning Right & Going Straight, Turning Left, Crossing Intersection, Approaching Intersection, Static, Parked. \\
\midrule
\multicolumn{2}{l}{\textit{Longitudinal Dynamics}} \\
\midrule
Speeding Up & Slowing Down, Moving Slowly, Static, Parked. \\
Slowing Down & Speeding Up, Moving Slowly, Static, Parked. \\
Varying Speed & Slowing Down, Speeding Up, Moving Slowly, Static, Parked. \\
Moving Slowly & Static, Parked. \\
\midrule
\multicolumn{2}{l}{\textit{Intersection Interaction}} \\
\midrule
Approaching Intersection & Crossing Intersection, Static, Parked, Turning Left, Turning Right, Speeding Up, Varying Speed. \\
Crossing Intersection & Approaching Intersection, Turning Left, Turning Right, Static, Parked. \\
\midrule
\multicolumn{2}{l}{\textit{Lane Position}} \\
\midrule
In Leftmost Lane & In Middle Lane, In Rightmost Lane, Static, Parked, all Lane Changes. \\
In Middle Lane & In Leftmost Lane, In Rightmost Lane, Static, Parked, all Lane Changes. \\
In Rightmost Lane & In Leftmost Lane, In Middle Lane, Static, Parked, all Lane Changes. \\
\midrule
\multicolumn{2}{l}{\textit{Lane Change Maneuvers}} \\
\midrule
All Lane Changes & In Any Lane (Left/Right/Mid), Static, Parked, and any disconnected/opposing Lane Changes (e.g., \textit{Change Left $\to$ Mid} is incompatible with \textit{Change Right $\to$ Mid}). \\
\bottomrule
\end{tabularx}
\vspace*{\fill}  
\end{table*}

\clearpage

\onecolumn
\begin{tcolorbox}[
    enhanced,
    breakable,
    colback=white,
    colframe=black,
    title={Orchestrator Prompt},
    fonttitle=\bfseries,
    width=\textwidth,
    boxrule=0.8pt
]

\small  

You are a professional visual editing assistant capable of performing object editing operations on videos based on user language instructions.

\tcbsubtitle{Available Functions:}

\textbf{Core Setup Functions:}

\begin{enumerate}[leftmargin=*,nosep,itemsep=2pt]
    
    \item \texttt{generate\_scene\_metadata(...)} - Generate the scene metadata
    
    \item \texttt{generate\_renderer\_kwargs()} - Generate the renderer configuration
    
    \item \texttt{create\_global\_simulator(...)} - Create the global simulator instance

    \item \texttt{create\_behavior\_simulator(...)} - Create the behavior simulator instance
\end{enumerate}

\vspace{0.2cm}
\textbf{Instruction Processing Function:}

\begin{enumerate}[leftmargin=*,nosep,itemsep=2pt,start=5]
    \item \texttt{rephrase\_instruction\_to\_id\_based(...)} - Rephrase the natural language instruction with object IDs (\textit{Object Grounding Agent})
\end{enumerate}

\vspace{0.2cm}
\textbf{Rendering Functions:}

\begin{enumerate}[leftmargin=*,nosep,itemsep=2pt,start=6]
    \item \texttt{render\_orig\_bev\_and\_rgb\_video(...)} - Render original BEV and RGB videos 

    \item \texttt{render\_behavior\_vector\_map(...)} - Render the object trajectories as a vector map video
    
    \item \texttt{render\_edited\_scene\_graph(...)} - Render the edited scene graph (\textit{Coarse Rendering Tool})
\end{enumerate}

\vspace{0.2cm}
\textbf{Object Editing Functions:}

\begin{enumerate}[leftmargin=*,nosep,itemsep=2pt,start=9]
    \item \texttt{remove\_object(...)} - Remove object nodes from the scene graph (\textit{Removal Tool})

    \item \texttt{retrieve\_from\_hunyuan(...)} - Generate a new 3D object mesh from the text description (\textit{Text-to-3D Tool})

    \item \texttt{rescale\_and\_transform\_mesh(...)} - Align the mesh's scale and local coordinate system with the scene, and add it to the scene graph (\textit{Insertion Agent})
    
    \item 
    \texttt{replace\_object(...)} - Replace the existing object node with a new one (\textit{Replacement Agent})

    \item \texttt{calculate\_initial\_position(...)} - Calculate the initial position for the new car

\end{enumerate}

\vspace{0.2cm}
\textbf{Behavior Simulation Functions:}

\begin{enumerate}[leftmargin=*,nosep,itemsep=2pt,start=14]
    \item \texttt{generate\_behavior\_description(...)} - Analyze the original trajectory to generate behavior descriptions

    \item \texttt{generate\_counterfactual\_behavior(...)} - Generate the counterfactual behavior combination list and select the best-matching (\textit{Behavior Editing Agent})
    
    \item \texttt{generate\_trajectory(...)} - Generate multi-object trajectories based on the selected counterfactual behavior combination (\textit{Multi-object Simulation Tool})
    
    \item \texttt{review\_and\_refine\_trajectories(...)} - Review and refine generated trajectories (\textit{Behavior Reviewer Agent})
\end{enumerate}

\vspace{0.2cm}
\textbf{Camera Control Functions:}

\begin{enumerate}[leftmargin=*,nosep,itemsep=2pt,start=18]
    \item \texttt{translate\_camera(...)} - Translate camera
    
    \item \texttt{rotate\_camera(...)} - Rotate camera
\end{enumerate}

\vspace{0.2cm}
\textbf{Video Refinement Function:}

\begin{enumerate}[leftmargin=*,nosep,itemsep=2pt,start=20]
    \item \texttt{refine\_with\_vdm(...)} - Generate refined video (\textit{Video Diffusion Tool})
\end{enumerate}

\tcbsubtitle{Standard Workflow Templates:}

All examples use the same standard code blocks. Only the \textbf{Editing Operations} section varies.

\textbf{Template A: Setup Phase}

\begin{tcolorbox}[
    colback=gray!10,
    colframe=gray!60,
    boxrule=0.5pt,
    breakable
]
\footnotesize\ttfamily
\# Generate metadata and create simulators\\
logging.info("Step 1: Generating metadata")\\
scene\_metadata = generate\_scene\_metadata(...)\\
renderer\_kwargs = generate\_renderer\_kwargs()\\
global\_simulator = create\_global\_simulator(...)\\
behavior\_simulator = create\_behavior\_simulator(...)\\
\\
\# Rephrase instruction and identify target object\\
logging.info("Step 2: Parsing the user instruction and identifying the target object node")\\
rephrased\_instruction = rephrase\_instruction\_to\_id\_based(...)\\
\\
\# Render original scene\\
logging.info("Step 3: Rendering the original scene")\\
bev\_orig\_path, video\_orig\_path = render\_orig\_bev\_and\_rgb\_video(...)\\
original\_trajectory\_vid\_path = render\_behavior\_vector\_map(...)\\
\\
\# Generate behavior description\\
logging.info("Step 4: Generating the behavior description")\\
generate\_behavior\_description(...)

\end{tcolorbox}

\textbf{Template B: Modified Scene Rendering}

\begin{tcolorbox}[
    colback=gray!10,
    colframe=gray!60,
    boxrule=0.5pt,
    breakable
]
\footnotesize\ttfamily
logging.info("Step 6: Rendering the modified scene")\\
modified\_trajectory\_vid\_path =  render\_behavior\_vector\_map(...)\\
coarse\_video\_path = render\_edited\_scene\_graph(...)
\end{tcolorbox}

\textbf{Template C: Video Refinement}

\begin{tcolorbox}[
    colback=gray!10,
    colframe=gray!60,
    boxrule=0.5pt,
    breakable
]
\footnotesize\ttfamily
logging.info("Step 7: Video Refinement")\\
refined\_video\_path = refine\_with\_vdm(...)
\end{tcolorbox}

\tcbsubtitle{Editing Operations Patterns:}

\textbf{1. Object Manipulation:}

Remove object:
\begin{tcolorbox}[colback=gray!10,colframe=gray!60,boxrule=0.5pt]
\footnotesize\ttfamily
logging.info("@@- Removing object")\\
remove\_object(...)
\end{tcolorbox}

Add new object:
\begin{tcolorbox}[colback=gray!10,colframe=gray!60,boxrule=0.5pt]
\footnotesize\ttfamily
target\_obj = retrieve\_from\_hunyuan(...)\\
\# IMPORTANT: Rescale and transform the generated mesh:\\
target\_obj = rescale\_and\_transform\_mesh(...)
\end{tcolorbox}

Replace with new object:
\begin{tcolorbox}[colback=gray!10,colframe=gray!60,boxrule=0.5pt]
\footnotesize\ttfamily
logging.info("@@- Replacing with new object")\\
new\_obj = replace\_object(...)
\end{tcolorbox}

\textbf{2. Trajectory/Behavior Generation:}

\begin{tcolorbox}[colback=gray!10,colframe=gray!60,boxrule=0.5pt]
\footnotesize\ttfamily
generate\_counterfactual\_behavior(...)\\
generate\_trajectory(...)\\
review\_and\_refine\_trajectories(...)
\end{tcolorbox}

\textbf{3. Camera Operations:}

\begin{tcolorbox}[colback=gray!10,colframe=gray!60,boxrule=0.5pt]
\footnotesize\ttfamily
translate\_camera(...)\\
rotate\_camera(...)
\end{tcolorbox}

\tcbsubtitle{Example:}

\textbf{Input}: ``Add a red sports car to the right of the yellow car and make it turn right."

\textbf{Output}: Template A + Core Editing + Template B + Template C

\textit{Core Editing Operation:}
\begin{tcolorbox}[colback=gray!10,colframe=gray!60,boxrule=0.5pt]
\footnotesize\ttfamily
logging.info("@@- Adding the new generated vehicle")\\
target\_obj = retrieve\_from\_hunyuan(...)\\
logging.info("@@@@• Aligning the mesh's scale and local coordinate system with the scene")\\
target\_obj = rescale\_and\_transform\_mesh(...)\\
\\
logging.info("@@- Calculating initial position")\\
initial\_position = calculate\_initial\_position(...)\\
\\
logging.info("@@- Generating counterfactual behavior, select the best-matching, generate trajectory and refine it")\\
generate\_counterfactual\_behavior(...)\\
generate\_trajectory(...)\\
review\_and\_refine\_trajectories(...)
\end{tcolorbox}

\captionof{figure}{Orchestrator prompt.}
\label{fig:orchestrator}
\end{tcolorbox}

\clearpage
\twocolumn

\onecolumn
\begin{tcolorbox}[
    enhanced,
    breakable,
    colback=white,
    colframe=black,
    title={Object Grounding Agent Prompt},
    fonttitle=\bfseries,
    width=\textwidth,
    boxrule=0.8pt
]

\small

Your task is to identify target objects in a driving scene graph based on textual descriptions.
Your goal is to find the objects that match the given description by analyzing their appearance, behavior and spatial information.

\tcbsubtitle{Task Overview:}

Given a textual description of an object (e.g., ``the red car on the left"), you need to:
\begin{enumerate}[nosep]
    \item Decompose the description into structured triplets
    \item Identify the reference object and filter candidates by direction
    \item Match attributes to find the target object
    \item Return the ID(s) of matching object(s)
\end{enumerate}

\tcbsubtitle{Step 1: Triplet Decomposition}

Extract natural-language descriptions of EXISTING objects that need ID conversion from the instruction.

\textbf{IMPORTANT RULES:}
\begin{enumerate}[nosep,leftmargin=*]
    \item IGNORE descriptions that already specify an ID (like ``car 2'', ``vehicle id 5'') - leave them unchanged in the final instruction.
    \item ONLY extract mentions of EXISTING objects that need ID conversion for operations like remove, replace, modify, etc.
    \item For ``add" operations: IGNORE the new objects being added, but DO extract any existing reference objects used to specify the new object's location.
    \item DO NOT extract the ego vehicle itself as an entity needing ID conversion.
    \begin{itemize}[nosep]
        \item Treat mentions like ``ego vehicle", ``ego car", ``camera car", ``our car" as the ego reference only; they should NOT appear in the returned list.
        \item If the instruction ONLY mentions the ego vehicle (e.g., lane change of the ego), return an empty list [].
        \item When ego is used as a spatial reference (e.g., ``the car on the left of the ego vehicle"), set reference\_desc to ``ego'' for that entity, but do not include the ego vehicle itself as an extracted entity.
    \end{itemize}
\end{enumerate}

\textbf{Output Format:}

For each EXISTING object mention that needs ID conversion, produce a JSON object with:

\begin{itemize}[nosep,leftmargin=*]
    \item \textbf{nl\_phrase}: exact substring from the original instruction that identifies the existing target object.
    \item \textbf{reference\_desc}: the object used as reference for location. Use ``ego'' if referring to the ego car, or a descriptive phrase if referring to another object. DEFAULT: ``ego'' when no reference is explicitly mentioned.
    \item \textbf{direction}: MUST be exactly one of [front, back, left, right] or null. Map all directional terms:
    \begin{itemize}[nosep]
        \item front: ahead, forward, in front of, fwd, etc.
        \item back: behind, rear, backward, etc.
        \item left: to the left, on the left side, etc.
        \item right: to the right, on the right side, etc.
        \item null: when no direction is specified (JSON null value).
    \end{itemize}
    \item \textbf{target\_desc}: descriptive attributes of the existing target object (color, type, brand, etc.). Use null if not described with specific attributes.
    \item \textbf{type}: MUST be exactly ``vehicle'' or ``pedestrian''. Determine based on the object description:
    \begin{itemize}[nosep]
        \item vehicle: cars, trucks, buses, vans, motorcycles, bicycles, etc.
        \item pedestrian: people, persons, humans, walkers, etc.
    \end{itemize}
\end{itemize}

\tcbsubtitle{Examples:}

\textbf{Example 1:}
\begin{tcolorbox}[colback=gray!10,colframe=gray!60,boxrule=0.5pt]
\footnotesize
\textbf{Input:} ``remove the red car"\\
\textbf{Output:} [\{``nl\_phrase": ``the red car", ``reference\_desc": ``ego", ``direction": null, ``target\_desc": ``red car", ``type": ``vehicle"\}]\\
\textbf{Explanation:} ``the red car" is an existing object being removed. No explicit reference mentioned, so reference is ``ego''. No direction specified, so direction is null.
\end{tcolorbox}

\textbf{Example 2:}
\begin{tcolorbox}[colback=gray!10,colframe=gray!60,boxrule=0.5pt]
\footnotesize
\textbf{Input:} ``add a blue pickup truck and remove the silver SUV in front"\\
\textbf{Output:} [\{``nl\_phrase": ``the silver SUV in front", ``reference\_desc": ``ego", ``direction": ``front", ``target\_desc": ``silver SUV", ``type": "vehicle"\}]\\
\end{tcolorbox}

\textbf{Example 3:}
\begin{tcolorbox}[colback=gray!10,colframe=gray!60,boxrule=0.5pt]
\footnotesize
\textbf{Input:} ``Have both the ego vehicle and the tan sedan on the left change to the middle lane"\\
\textbf{Output:} [\{``nl\_phrase": ``the tan sedan on the left", ``reference\_desc": ``ego", ``direction": ``left", ``target\_desc": ``tan sedan", ``type": ``vehicle"\}]\\
\textbf{Explanation:} The ego vehicle is NOT extracted as an entity needing ID conversion. Only the tan sedan is extracted, with default reference as ``ego'' and direction ``left''.
\end{tcolorbox}

\vspace{0.3cm}
\textbf{Instruction:} ``\{instruction\}"

\textbf{Return ONLY the JSON array.}

\tcbsubtitle{Step 2\&3\&4: Object Grounding}

After extracting triplets in Step 1, you will use multi-modal information to identify the target object(s). The input consists of:
\begin{itemize}[nosep]
    \item \textbf{reference\_desc}: Description of the reference object.
    \item \textbf{direction}: Spatial direction relative to reference (front, back, left, right, or null).
    \item \textbf{target\_desc}: Description of the target object.
\end{itemize}

\textbf{Scene Information Provided:}

You will receive the following information about all objects in the scene:

\begin{enumerate}[nosep,leftmargin=*]
    \item \textbf{Appearance (Visual)}:
    \begin{itemize}[nosep]
        \item An image of the driving scene from the ego vehicle's dash cam perspective.
        \item Each object's center is labeled with its ID number in red text.
        \item The ego vehicle is NOT visible (it is taking the photo).
        \item ONLY objects with visible ID numbers should be considered.
    \end{itemize}
    
    \item \textbf{Behavior (Textual)}:
    \begin{itemize}[nosep]
        \item Behavior description for each object.
        \item Describes trajectory motion and lane information.
        \item Examples: ``in the leftmost lane, speed up, go straight".
    \end{itemize}
    
    \item \textbf{Position Information}:
    \begin{itemize}[nosep]
        \item Complete trajectory coordinates for each object.
        \item Coordinates follow the Waymo world coordinate system:
        \begin{itemize}[nosep]
            \item \textbf{X-axis}: Forward direction (vehicle's front).
            \item \textbf{Y-axis}: Left direction (vehicle's left side).
            \item \textbf{Z-axis}: Upward direction (vertical).
        \end{itemize}
        \item Map information including:
        \begin{itemize}[nosep]
            \item Lane topology (predecessor and successor lanes).
            \item Left and right neighbor lane information for each lane.
        \end{itemize}
    \end{itemize}
\end{enumerate}

\textbf{Three-Step Matching Process:}

\begin{enumerate}[nosep,leftmargin=*]
    \item \textbf{Find Reference Object:}
    \begin{itemize}[nosep]
        \item If reference\_desc is ``ego", use the ego vehicle as reference
        \item Otherwise, match reference\_desc against all objects using:
        \begin{itemize}[nosep]
            \item Appearance: color, type, size (from image).
            \item Behavior: motion pattern, lane position (from behavior description).
            \item Position: spatial location (from trajectory coordinates and map).
        \end{itemize}
        \item Select the object ID that best matches reference\_desc.
    \end{itemize}
    
    \item \textbf{Filter Candidates by Direction:}
    \begin{itemize}[nosep]
        \item If direction is null, consider all objects as candidates.
        \item Otherwise, filter objects based on the specified direction:
        \begin{itemize}[nosep]
            \item Use trajectory coordinates and map lane information to determine spatial relationships.
            \item Apply direction constraints (front, back, left, right) relative to reference object.
            \item Consider strict\_direction flag for front/back filtering if applicable.
            \item For left/right: ensure candidates are in different lanes from reference.
        \end{itemize}
        \item Form a candidate set containing only objects in the specified direction.
    \end{itemize}
    
    \item \textbf{Find Target Object in Candidates:}
    \begin{itemize}[nosep]
        \item From the filtered candidate set, match target\_desc using the same multi-modal approach:
        \begin{itemize}[nosep]
            \item Appearance matching from image.
            \item Behavior matching from behavior descriptions.
            \item Position verification from trajectory coordinates and map information.
        \end{itemize}
        \item Return the object ID(s) that best match target\_desc.
        \item If multiple objects match equally well, return the nearest one.
    \end{itemize}
\end{enumerate}

\captionof{figure}{Object grounding agent prompt.}
\label{fig:object_grounding_agent}
\end{tcolorbox}

\clearpage
\twocolumn

\onecolumn
\begin{tcolorbox}[
    enhanced,
    breakable,
    colback=white,
    colframe=black,
    title={Insertion Agent Prompt},
    fonttitle=\bfseries,
    width=\textwidth,
    boxrule=0.8pt
]

\small

Your task is to prepare generated 3D vehicle meshes for insertion into a driving scene. Your goal is to: (1) calculate the scaling factor to resize the mesh to real-world size, and (2) compute the transformation matrix from local to world coordinates.

\tcbsubtitle{Input Information:}

You will receive the following information:

\begin{itemize}[nosep,leftmargin=*]
    \item \textbf{Text Description}: A natural language description of the vehicle (e.g., ``blue sedan'', ``red sports car'')
    \item \textbf{Generated Mesh Bounding Box}: The 3D bounding box dimensions of the generated mesh.
    \item \textbf{Rendered Images}: Images of the mesh rendered along specified axes.
    \item \textbf{World Coordinate System}: The scene's world coordinate system follows Waymo convention:
    \begin{itemize}[nosep]
        \item \textbf{X-axis}: Forward direction (vehicle's front).
        \item \textbf{Y-axis}: Left direction (vehicle's left side).
        \item \textbf{Z-axis}: Upward direction (vertical).
    \end{itemize}
\end{itemize}

\tcbsubtitle{Step 1: Calculate Scaling Factor}

You are a vehicle dimensions expert. Given a vehicle description, provide the typical real-world dimensions for that specific vehicle in meters.

Vehicle description: ``\{description\}''

Mesh bounding box dimensions: \{mesh\_bbox\}

Please analyze the description and provide realistic dimensions for this specific vehicle. Consider:
\begin{itemize}[nosep]
    \item The exact vehicle type mentioned (if specific model/brand is mentioned, use those dimensions).
    \item Typical size ranges for that category of vehicle.
    \item Any size indicators in the description (compact, large, etc.).
\end{itemize}

After determining the real-world dimensions, calculate the scaling factor using:

scaling\_factor = real\_world\_dimension / mesh\_bounding\_box\_dimension

Apply appropriate scaling factors for each dimension (length, width, height) to resize the mesh to real-world scale.

\tcbsubtitle{Step 2: Compute Transformation Matrix}

Analyze the heading direction of the vehicle in the rendered image, then compute the transformation matrix from local to world coordinates.

\textbf{Part 2.1: Analyze Vehicle Heading Direction}

Analyze the heading direction of the vehicle in the image. Please provide your reasoning process first, then give your final answer.

\textbf{Direction definitions:}
\begin{itemize}[nosep]
    \item \textbf{forward}: Vehicle front is facing toward the camera/viewer.
    \item \textbf{backward}: Vehicle rear is facing toward the camera/viewer.
    \item \textbf{left}: Vehicle front is pointing to the left side of the image.
    \item \textbf{right}: Vehicle front is pointing to the right side of the image.
\end{itemize}

The answer should be one of these four options: forward, backward, left, right.

Please follow this format:
\begin{enumerate}[nosep]
    \item First, describe what you observe about the vehicle's orientation and features
    \item Explain your reasoning for determining the heading direction
    \item On the final line, write only one word: forward, backward, left, or right
\end{enumerate}

\textbf{Part 2.2: Determine Local Coordinate System and Compute Transformation}

Based on the identified heading direction, the local coordinate system is defined as follows:

\begin{itemize}[nosep,leftmargin=*]
    \item \textbf{forward}: car\_head\_direction: +z, length\_axis: z, width\_axis: x, height\_axis: y.\\
    Description: Car head points in +z direction.
    
    \item \textbf{backward}: car\_head\_direction: -z, length\_axis: z, width\_axis: x, height\_axis: y.\\
    Description: Car head points in -z direction.
    
    \item \textbf{left}: car\_head\_direction: -x, length\_axis: x, width\_axis: z, height\_axis: y.\\
    Description: Car head points in -x direction.
    
    \item \textbf{right}: car\_head\_direction: +x, length\_axis: x, width\_axis: z, height\_axis: y.\\
    Description: Car head points in +x direction.
\end{itemize}

Using the local coordinate system definition and the world coordinate system (X: forward, Y: left, Z: up), compute the transformation matrix that converts coordinates from the local mesh coordinate system to the world coordinate system.

The transformation matrix should be a 4×4 matrix in homogeneous coordinates.

\textbf{Output Format:}

Respond with ONLY a JSON object in this exact format:
\begin{verbatim}
{
    "vehicle_type": "brief description of vehicle type",
    "real_world_dimensions": {
        "length": X.X,
        "width": X.X,
        "height": X.X
    },
    "scaling_factor": {
        "length": X.X,
        "width": X.X,
        "height": X.X
    },
    "heading_direction": "forward/backward/left/right",
    "transformation_matrix": [
        [X, X, X, X],
        [X, X, X, X],
        [X, X, X, X],
        [X, X, X, X]
    ]
}
\end{verbatim}

Where all dimensions are in meters as decimal numbers, and the transformation matrix is a 4×4 matrix.

Do not include any other text or explanation beyond the JSON object.

\captionof{figure}{Insertion agent prompt for mesh scaling and coordinate transformation.}
\label{fig:insertion_agent}
\end{tcolorbox}

\clearpage
\twocolumn

\onecolumn
\begin{tcolorbox}[
    enhanced,
    breakable,
    colback=white,
    colframe=black,
    title={Behavior Editing Agent Prompt},
    fonttitle=\bfseries,
    width=\textwidth,
    boxrule=0.8pt
]

\small

Your task is to select appropriate counterfactual behavior combination for objects in a driving scene. 
Your goal is to match the target behavior from user instructions with available counterfactual behavior combination list while preserving as much of the original behaviors as possible.

\tcbsubtitle{Input Information:}

You will receive the following information for each object:

\begin{itemize}[nosep,leftmargin=*]
    \item \textbf{Target Behavior}: The desired behavior extracted from the user instruction (with object IDs).
    \item \textbf{Original Behavior}: The object's original behavior before modification.
    \item \textbf{Available Counterfactuals}: A list of alternative behavior combinations for each object.
\end{itemize}

\tcbsubtitle{Task:}

\begin{enumerate}[nosep,leftmargin=*]    
    \item \textbf{For each object, find counterfactuals that COMPLETELY CONTAIN the requested behavior:}
    \begin{itemize}[nosep]
        \item The counterfactual MUST include every single behavior mentioned in the request with identical meaning.
        \item STRICT MATCHING ONLY: the counterfactual behavior must contain behaviors with exactly the same meaning as the requested behaviors.
        \item When multiple counterfactuals contain all requested behaviors, prioritize: smallest differences from that object's original behavior.
        \item If an object has NO counterfactual that completely contains all the requested behaviors, return ``none" for that object.
    \end{itemize}
\end{enumerate}

\tcbsubtitle{Matching Rules:}

\begin{enumerate}[nosep,leftmargin=*]
    \item \textbf{STRICT MATCHING REQUIRED:} Counterfactual behaviors must completely contain ALL requested behaviors with identical meaning.

    \item \textbf{SELECTION PRIORITY:} When multiple counterfactuals match: use smallest differences from that object's original behavior.
    
    \item \textbf{OUTPUT FORMAT:} Use the EXACT text from the selected counterfactual, not your own interpretation.
    
    \item \textbf{NO MATCH CASE:} If NO counterfactual completely contains the requested behavior for any object, output ``none".
    
    \item \textbf{MULTI-OBJECT:} For multi-object behaviors, match each object's part with their respective counterfactuals using the same strict rules.
\end{enumerate}

\tcbsubtitle{Output Format:}

Each object should be on a separate line in the format: ``object\_id: counterfactual\_behavior".

\begin{itemize}[nosep,leftmargin=*]
    \item Use the original ID format (examples: ``car [number]", ``pedestrian [number]", ``cyclist [number]", ``ego", etc.).
    \item If a counterfactual matches, return that counterfactual behavior.
    \begin{itemize}[nosep]
        \item Example: ``car 123: going straight, in middle lane, crossing an intersection".
    \end{itemize}
    \item If no counterfactual matches for an object, output ``none".
    \begin{itemize}[nosep]
        \item Example: ``car 123: none".
    \end{itemize}
\end{itemize}

\tcbsubtitle{Examples:}

\textbf{Example 1: Simple lane change}
\begin{tcolorbox}[colback=gray!10,colframe=gray!60,boxrule=0.5pt]
\footnotesize
\textbf{Target behavior:} ``car 2 changes lanes from middle lane to rightmost lane''\\
\textbf{Current behaviors before modification:}\\
\phantom{xx}- 2: ``going straight, in middle lane, crossing an intersection''\\
\textbf{Available counterfactuals for 2:}\\
\phantom{xx}[``going straight, in middle lane, crossing an intersection'',\\
\phantom{xxx}``going straight, changing lanes from middle lane to rightmost lane, crossing an intersection'']\\
\\
\textbf{Output:}\\
car 2: going straight, changing lanes from middle lane to rightmost lane, crossing an intersection\\
\\
\textbf{Explanation:} Matches the second counterfactual because it contains ``changing lanes from middle lane to rightmost lane"
\end{tcolorbox}

\textbf{Example 2: No matching counterfactual}
\begin{tcolorbox}[colback=gray!10,colframe=gray!60,boxrule=0.5pt]
\footnotesize
\textbf{Target behavior:} ``car 2 turns left''\\[0.2cm]
\textbf{Current behaviors before modification:}\\
\phantom{xx}- 2: ``going straight, in leftmost lane, crossing an intersection''\\[0.2cm]
\textbf{Available counterfactuals for 2:}\\
\phantom{xx}[``going straight, in leftmost lane, crossing an intersection'',\\
\phantom{xxx}``going straight, speeding up'']\\[0.2cm]
\textbf{Output:}\\
car 2: none\\[0.2cm]
\textbf{Explanation:} No match because none of the counterfactuals contain ``turning left'' - they only have ``going straight''
\end{tcolorbox}

\textbf{Example 3: Multi-object behavior}
\begin{tcolorbox}[colback=gray!10,colframe=gray!60,boxrule=0.5pt]
\footnotesize
\textbf{Target behavior:} ``car 2 speeds up while car 5 turns left''\\[0.2cm]
\textbf{Current behaviors before modification:}\\
\phantom{xx}- 2: ``going straight, in rightmost lane, crossing an intersection''\\
\phantom{xx}- 5: ``going straight, in middle lane, crossing an intersection''\\[0.2cm]
\textbf{Available counterfactuals for 2:}\\
\phantom{xx}[``going straight, in rightmost lane, crossing an intersection'',\\
\phantom{xxx}``going straight, speeding up, in rightmost lane, crossing an intersection'']\\[0.2cm]
\textbf{Available counterfactuals for 5:}\\
\phantom{xx}[``going straight, speeding up, in middle lane, crossing an intersection'',\\
\phantom{xxx}``going straight, changing lanes from middle lane to leftmost lane, crossing an intersection'']\\[0.2cm]
\textbf{Output:}\\
car 2: going straight, speeding up, in rightmost lane, crossing an intersection\\
car 5: none\\[0.2cm]
\textbf{Explanation:} car 2 matches because counterfactual contains ``speeding up'', car 5 has no match because no counterfactual contains ``turning left''
\end{tcolorbox}

\captionof{figure}{Behavior editing agent prompt for selecting counterfactual behaviors.}
\label{fig:behavior_editing_agent}
\end{tcolorbox}

\clearpage
\twocolumn

\onecolumn
\begin{tcolorbox}[
    enhanced,
    breakable,
    colback=white,
    colframe=black,
    title={Behavior Reviewer Agent Prompt},
    fonttitle=\bfseries,
    width=\textwidth,
    boxrule=0.8pt
]

\small

Your task is to review generated trajectories for objects in a driving scene and adjust guidance configurations to improve trajectory realism. 
Your goal is to analyze evaluation results and determine appropriate guidance mode and weight adjustments for each object.

\tcbsubtitle{Input Information:}

You will receive the following information for each object:

\begin{itemize}[nosep,leftmargin=*]
    \item \textbf{Validation Results}: Assessment of the generated trajectory across three aspects:
    \begin{itemize}[nosep]
        \item \textbf{Behavior Alignment}: Whether the trajectory matches the target behavior.
        \item \textbf{On-Road}: Whether the trajectory stays on valid road areas.
        \item \textbf{No Collision}: Whether the trajectory avoids collisions with other objects.
    \end{itemize}
    
    \item \textbf{Current Mode}: The trajectory generation mode used for this object:
    \begin{itemize}[nosep]
        \item \textbf{cf\_guidance}: Trajectory generated using textual description as condition (classifier-free guidance).
        \item \textbf{pre\_traj\_guidance}: Trajectory generated using previously successful trajectory as guidance.
    \end{itemize}
    
    \item \textbf{Current Guidance Configuration}: Active guidances and their weights:
    \begin{itemize}[nosep]
        \item In \textbf{cf\_guidance mode}: Initially only classifier-free guidance is active; on-road and no-collision guidances can be added.
        \item In \textbf{pre\_traj\_guidance mode}: Only pre-traj guidance is active.
    \end{itemize}
\end{itemize}

\tcbsubtitle{Task:}

For each object, analyze its evaluation results and adjust the guidance configuration according to the following rules:

\textbf{Case 1: Object in cf\_guidance mode}

Available guidances in this mode:
\begin{itemize}[nosep]
    \item classifier-free guidance (always present).
    \item on-road guidance (added when needed).
    \item no-collision guidance (added when needed).
\end{itemize}

\textbf{If trajectory satisfies ALL three conditions} (behavior alignment, on-road, and no collision):
\begin{itemize}[nosep]
    \item Switch mode to \textbf{pre\_traj\_guidance}.
    \item Save the current successful trajectory as guidance.
    \item Replace all guidances with only pre-traj guidance (initial weight e.g., 1e4).
\end{itemize}

\textbf{If trajectory fails ANY condition:}

\begin{itemize}[nosep]
    \item \textbf{Behavior alignment failed}:
    \begin{itemize}[nosep]
        \item Increase classifier-free guidance weight by 1.0
        \item Formula: new\_weight = current\_weight + 1.0
    \end{itemize}
    
    \item \textbf{On-road failed}:
    \begin{itemize}[nosep]
        \item If on-road guidance does NOT exist: add it with initial weight 1e3
        \item If on-road guidance already exists: multiply its weight by 3.0
        \item Formula: new\_weight = current\_weight × 3.0
    \end{itemize}
    
    \item \textbf{No collision failed}:
    \begin{itemize}[nosep]
        \item If no-collision guidance does NOT exist: add it with initial weight 1e3
        \item If no-collision guidance already exists: multiply its weight by 3.0
        \item Formula: new\_weight = current\_weight × 3.0
    \end{itemize}
\end{itemize}

\textbf{Note:} If multiple conditions fail, apply ALL corresponding adjustments. Stay in cf\_guidance mode.

\textbf{Case 2: Object in pre\_traj\_guidance mode}

Available guidance in this mode:
\begin{itemize}[nosep]
    \item pre-traj guidance (only guidance in this mode).
\end{itemize}

\textbf{If trajectory is successful} (satisfies all three conditions):
\begin{itemize}[nosep]
    \item Maintain current mode (pre\_traj\_guidance).
    \item Keep pre-traj guidance weight unchanged.
\end{itemize}

\textbf{If trajectory fails} (any condition not satisfied):
\begin{itemize}[nosep]
    \item Increase pre-traj guidance weight by multiplying by 3.0
    \item Formula: new\_weight = current\_weight × 3.0
    \item Stay in pre\_traj\_guidance mode.
\end{itemize}

\tcbsubtitle{Output Format:}

For each object, provide the updated configuration in the following JSON format:

\begin{verbatim}
{
    "object_id": {
        "mode": "cf_guidance" or "pre_traj_guidance",
        "guidance_config": {
            // For cf_guidance mode:
            "classifier_free": weight_value,
            "on_road": weight_value (if added),
            "no_collision": weight_value (if added)
            
            // For pre_traj_guidance mode:
            "pre_traj": weight_value
        },
        "status": "success" or "failed",
        "failed_aspects": ["aspect1", "aspect2", ...] (if failed),
        "adjustments_made": ["description of adjustments"]
    }
}
\end{verbatim}

Where:
\begin{itemize}[nosep]
    \item \textbf{mode}: Current generation mode after adjustment.
    \item \textbf{guidance\_config}: Dictionary of active guidances and their weights (format depends on mode).
    \item \textbf{status}: Whether the trajectory evaluation was successful.
    \item \textbf{failed\_aspects}: List of which aspects failed (if any).
    \item \textbf{adjustments\_made}: Description of what changes were made.
\end{itemize}

\textbf{Important:}
\begin{itemize}[nosep]
    \item Return configurations for ALL objects in the scene.
    \item In cf\_guidance mode: only include classifier-free, on-road, and no-collision guidances.
    \item In pre\_traj\_guidance mode: only include pre-traj guidance.
    \item Only include guidances that are actually active (weight $>$ 0).
    \item Provide clear reasoning for each adjustment.
    \item Respond with ONLY the JSON object, no additional text.
\end{itemize}

\captionof{figure}{Behavior reviewer agent prompt.}
\label{fig:behavior_reviewer_agent}
\end{tcolorbox}

\clearpage

\onecolumn
\begin{tcolorbox}[
    enhanced,
    breakable,
    colback=white,
    colframe=black,
    title={Video Reviewer Agent Prompt},
    fonttitle=\bfseries,
    width=\textwidth,
    boxrule=0.8pt
]

\small

You are a professional video refinement reviewer. Your task is to analyze video frames produced by a video diffusion model (VDM) and adjust two hyperparameters to improve photorealism while preserving the inserted vehicle's appearance.

\tcbsubtitle{Goal:}

Given a coarse composited video frame (a Gaussian-Splatting-rendered background with a depth-composited 3D vehicle mesh) and the refined video frame generated by VDM, you will:
\begin{itemize}[nosep,leftmargin=*]
    \item Improve photorealism of the inserted vehicle, especially lighting consistency with the environment.
    \item Preserve the inserted vehicle's appearance from the coarse video frame (shape, key parts, vehicle type, color, etc.).
\end{itemize}

\tcbsubtitle{Inputs:}

You will receive:
\begin{itemize}[nosep,leftmargin=*]
    \item \textbf{Coarse video frame(s)}: VDM condition frames (Gaussian Splatting background + depth-composited 3D mesh vehicle).
    \item \textbf{Refined video frame(s)}: VDM output frames.
    \item \textbf{Vehicle mask(s)}: Binary masks for the inserted vehicle region in each video frame.
    \item \textbf{Current diffusion strength} \texttt{strength} and its upper bound \texttt{strength\_ub}.
    \item \textbf{Current L2 guidance loss weight} \texttt{l2\_weight} and its upper bound \texttt{l2\_weight\_ub}.
\end{itemize}

\tcbsubtitle{Task:}

For each provided \textbf{video-frame pair} (coarse vs. refined):
\begin{enumerate}[nosep,leftmargin=*]
    \item Use the mask to focus on the inserted vehicle region.
    \item Answer \textbf{only two questions} based on the masked-region comparison:
    \begin{enumerate}[nosep,leftmargin=*]
        \item \textbf{Realism \& lighting:} Does the refined inserted vehicle look realistic (i.e., the lighting matches the environment and there are no obvious artifacts)?
        \item \textbf{Appearance preservation:} Does the refined result preserve the coarse vehicle's appearance (shape, key parts, type, color, etc.)?
    \end{enumerate}
    \item Update hyperparameters according to the rules below.
\end{enumerate}

\tcbsubtitle{Update Rules:}

\textbf{Rule 1 (Realism \& lighting $\rightarrow$ strength).}
\begin{itemize}[nosep,leftmargin=*]
    \item If the answer to \textbf{Q1} is \textbf{NO}, increase diffusion strength:
    \[
        \texttt{strength\_new} = \frac{\texttt{strength} + \texttt{strength\_ub}}{2}.
    \]
    \item Otherwise, keep \texttt{strength} unchanged (\texttt{strength\_new} = \texttt{strength}).
    \item \textbf{Upper-bound constraint:} The maximum allowed value is \texttt{strength\_ub}. If the computed \texttt{strength\_new} exceeds \texttt{strength\_ub}, set \texttt{strength\_new} = \texttt{strength\_ub}.
\end{itemize}

\textbf{Rule 2 (Appearance preservation $\rightarrow$ L2 weight).}
\begin{itemize}[nosep,leftmargin=*]
    \item If the answer to \textbf{Q2} is \textbf{NO}, increase L2 guidance loss weight:
    \[
        \texttt{l2\_weight\_new} = \frac{\texttt{l2\_weight} + \texttt{l2\_weight\_ub}}{2}.
    \]
    \item Otherwise, keep \texttt{l2\_weight} unchanged (\texttt{l2\_weight\_new} = \texttt{l2\_weight}).
    \item \textbf{Upper-bound constraint:} The maximum allowed value is \texttt{l2\_weight\_ub}. If the computed \texttt{l2\_weight\_new} exceeds \texttt{l2\_weight\_ub}, set \texttt{l2\_weight\_new} = \texttt{l2\_weight\_ub}.
\end{itemize}

\textbf{Note:} Both rules may trigger simultaneously.

\tcbsubtitle{Output Format:}

Return ONLY a JSON object in the following format (no extra text):

\begin{verbatim}
{
  "q1_realism_and_lighting_ok": true/false,
  "q2_appearance_preserved": true/false,
  "update": {
    "strength_old": X.XXXX,
    "strength_ub": X.XXXX,
    "strength_new": X.XXXX,
    "l2_weight_old": X.XXXX,
    "l2_weight_ub": X.XXXX,
    "l2_weight_new": X.XXXX
  },
  "notes": {
    "q1_reason": "one short phrase",
    "q2_reason": "one short phrase"
  }
}
\end{verbatim}

\textbf{Important Constraints:}
\begin{itemize}[nosep,leftmargin=*]
    \item Base both answers strictly on the masked vehicle region in the video frames.
    \item Keep updates strictly according to the averaging rule with the provided upper bounds.
    \item After computing an update by averaging, \textbf{clamp} it to the upper bound if necessary (i.e., the final value must not exceed its \texttt{*\_ub}).
    \item If no update is needed for a parameter, set \texttt{*\_new} equal to \texttt{*\_old}.
    \item Keep reasons short and concrete (e.g., ``lighting too warm vs. background''; ``color shifted from green to black'').
\end{itemize}

\captionof{figure}{Video reviewer agent prompt.}
\label{fig:video_reviewer_agent}
\end{tcolorbox}

\end{document}